\documentclass[lettersize,journal]{IEEEtran}
\usepackage{xcolor} 
\usepackage[noadjust]{cite}

\usepackage{graphicx} 
\usepackage{comment}
\usepackage[noend]{algpseudocode} 

\usepackage[ruled,linesnumbered]{algorithm2e}
\usepackage{amsmath}
\usepackage{amssymb} 
\usepackage{amsthm,thmtools} 
\usepackage{mathtools} 
\usepackage{mathrsfs} 
\usepackage{fdsymbol} 
\usepackage{multirow}
\usepackage{indentfirst}
\usepackage{booktabs}
\usepackage{tabularx}
\usepackage{threeparttable}
\usepackage[font=small]{caption}
\usepackage[font=small]{subcaption}

\usepackage{color}
\usepackage{hyperref}
\usepackage{kotex}
\usepackage{diagbox}
\usepackage{makecell}

\newtheorem{theorem}{Theorem}

\newtheorem{lemma}{Lemma}

\theoremstyle{remark}

\newcolumntype{s}{>{\hsize=.6\hsize\linewidth=\hsize}X}
\newcolumntype{d}{>{\hsize=1.4\hsize\linewidth=\hsize}X}

\usepackage{pifont}
\newcommand{\cmarkbold}{\ding{52}}%
\newcommand{\xmark}{\ding{53}}%

\begin{document}

\title{MC-Swarm: Minimal-Communication Multi-Agent Trajectory Planning and Deadlock Resolution for Quadrotor Swarm}

\author{Yunwoo Lee$^{1}$ and Jungwon Park$^{2*}$
\thanks{$^{1}$The author is with AI Institute of Seoul National University, Seoul, South Korea, and Carnegie Mellon University, Pittsburgh, PA, USA (e-mail: yunwoo333@gmail.com) }
\thanks{$^{2}$The author is with the Department of Mechanical System Design Engineering, Seoul National University of Science and Technology (SEOULTECH), Seoul, South Korea (e-mail: jungwonpark@seoultech.ac.kr)}
\thanks{$^{*}$: corresponding author}}

\markboth{ }%
{Shell \MakeLowercase{\textit{et al.}}: A Sample Article Using IEEEtran.cls for IEEE Journals}


\maketitle

\begin{abstract} For effective multi-agent trajectory planning, it is important to consider lightweight communication and its potential asynchrony. This paper presents a distributed trajectory planning algorithm for a quadrotor swarm that operates asynchronously and requires no communication except during the initial planning phase. Moreover, our algorithm guarantees no deadlock under asynchronous updates and absence of communication during flight. To effectively ensure these points, we build two main modules: coordination state updater and trajectory optimizer. The coordination state updater computes waypoints for each agent toward its goal and performs subgoal optimization while considering deadlocks, as well as safety constraints with respect to neighbor agents and obstacles. Then, the trajectory optimizer generates a trajectory that ensures collision avoidance even with the asynchronous planning updates of neighboring agents. We provide a theoretical guarantee of collision avoidance with deadlock resolution and evaluate the effectiveness of our method in complex simulation environments, including random forests and narrow-gap mazes. Additionally, to reduce the total mission time, we design a faster coordination state update using lightweight communication.  Lastly, our approach is validated through extensive simulations and real-world experiments with cluttered environment scenarios. 
\end{abstract}

\begin{IEEEkeywords}
Path Planning for Multiple Mobile Robots, Collision Avoidance, Distributed Robot Systems.
\end{IEEEkeywords}

\section{Introduction}
\IEEEPARstart{T}{he} compactness of quadrotor drones enables the operation of multi-agent systems in cluttered environments. While small teams of drones can be manually controlled by human pilots, large-scale swarms require autonomous coordination, where multi-agent trajectory planning (MATP) serves as a critical component. Over the past decade, MATP has been extensively studied, leading to its adoption in various applications, such as surveillance \cite{tie_surveillance}, inspection \cite{multi_inspection}, and transportation \cite{multi_transportation}.

Many existing MATP frameworks rely on synchronous coordination, where agents repeatedly exchange information to maintain consistency during planning and execution \cite{intro_synch}. However, as the number of agents increases, the communication load grows significantly, often resulting in message delays and packet losses. These issues degrade overall performance, as agents may operate based on inconsistent or outdated information. Without a robust and carefully engineered communication infrastructure, maintaining synchronization becomes increasingly impractical. To address these limitations, recent studies have explored asynchronous MATP, which relaxes communication requirements by allowing agents to operate with limited or delayed information \cite{kondo2023robust}. However, most existing approaches still depend on partial data exchange (e.g., previous planning results), leaving them vulnerable to the same communication limitations they aim to overcome.

In addition to the communication constraints, deadlock resolution is another critical factor in ensuring the successful completion of swarm tasks. Deadlocks can occur not only from the increased number of agents but also in confined environments, such as warehouses, where agents may block one another's paths and fail to reach their goals. While several methods have been proposed to resolve deadlocks, many either lack theoretical guarantees \cite{abdullhak2021deadlock_notguarantee} or rely on synchronous updates \cite{chen2024deadlock_guarantee}, making them unsuitable for large-scale \cite{large_scale} or communication-constrained settings \cite{comm_restricted}.

\begin{figure}[t!]
\centering
\includegraphics[width = 0.9\linewidth]{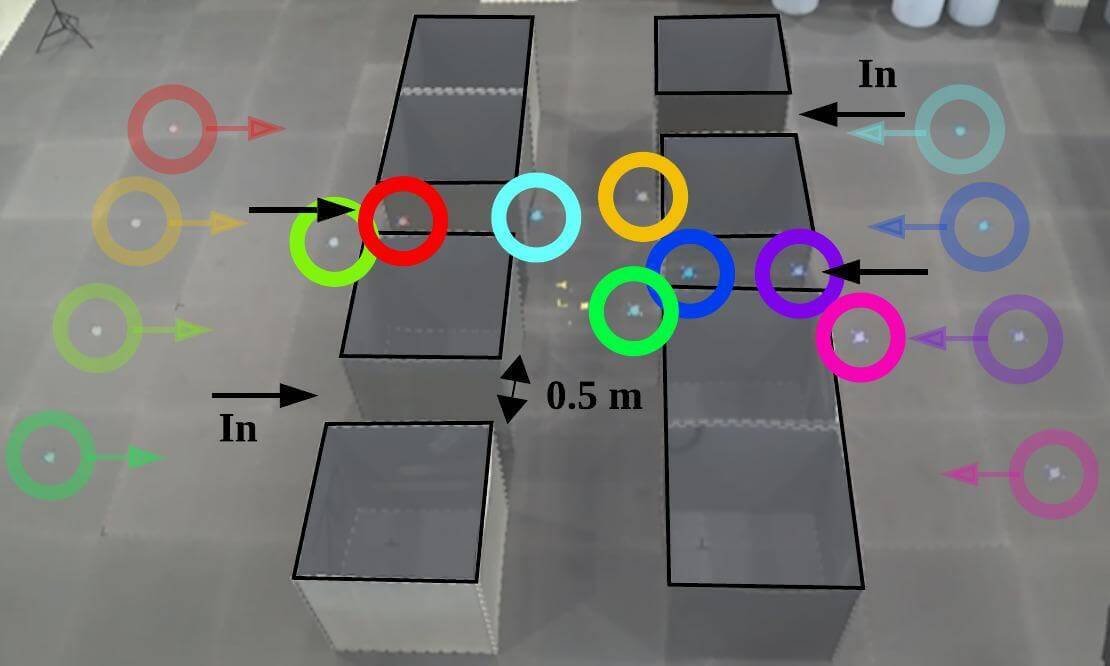}
\caption{Demonstration of a goal-reaching mission with eight quadrotors in a narrow-gap environment.}
\label{fig1:thumbnail}
\end{figure}

In this paper, we present MC-Swarm, an asynchronous and distributed MATP method that ensures deadlock resolution with minimal communication. The proposed framework consists of two main components: coordination state update and trajectory optimization.
During the coordination state update, each agent determines its coordination state, which includes waypoint, subgoal, and collision constraints, while considering deadlock resolution. Here, a waypoint is an intermediate target on the path to the final goal, and a subgoal is a local target that guides the agent toward the waypoint. Based on this coordination state, each agent then independently optimizes its trajectory toward the goal.
Our update policy ensures consistency in subgoal and waypoint information among agents, resolves deadlocks, and enables collision avoidance with obstacles and neighboring agents, despite asynchronous operation and lack of communication.
Then, the trajectory of each agent is optimized towards its goal, based on the calculated subgoals and safety constraints. 
To accommodate varying communication constraints, we introduce two variants of the method: a no-communication version (\textbf{MC-Swarm-N}) and a lightweight communication version (\textbf{MC-Swarm-C}). In the communication-free variant, agents exchange only their start and goal positions during the initial planning phase; no further communication is required during the flight. In the lightweight communication variant, limited information is shared to accelerate the coordination process and reduce the overall mission time.
The proposed method provides theoretical guarantees of deadlock resolution and collision avoidance. The proposed method is particularly well-suited for goal-reaching tasks, and its effectiveness is validated through extensive comparisons with state-of-the-art (SOTA) approaches and real-world experiments, as shown in Fig. \ref{fig1:thumbnail}.

Table \ref{table:comparison_with_sota} summarizes the comparison with the SOTA algorithms, and the main contributions of this paper are as follows:
\begin{itemize}
\item An asynchronous and distributed multi-agent trajectory planning (MATP) strategy that eliminates the need for communication after the initial planning phase.
\item A subgoal optimization method that guarantees deadlock resolution without communication, demonstrating superior performance in cluttered or narrow environments where SOTA algorithms often fail.
\item A trajectory optimization method that guarantees collision avoidance under asynchronous and communication-free updates by individual agents.
\item A coordination strategy that reduces total mission time by enabling limited communication.
\end{itemize}

The structure of the remainder of this article is as follows. Section \ref{sec2:related_work} reviews the related literature, and the problem formulation is presented in Section \ref{sec3:problem_formulation}. Then, Section \ref{sec4:coordination_state_update} explains the methods of waypoint updates, subgoal optimization, and collision constraints.
In Section \ref{sec5:trajectory_planning}, we present trajectory optimization to make drones reach subgoals while guaranteeing the drone's safety. Section \ref{sec6:validation} describes the comparison tests with SOTA algorithms through simulations and validates the proposed method through hardware demonstration. Finally, Section \ref{sec7:conclusion} concludes this paper.

\section{Related work}
\label{sec2:related_work}
\subsection{Communication-Aware MATP}
\label{subsec2:communication_aware_related}
Trajectory planning for multirotor UAV swarms has been extensively studied over the past decade. Early works primarily adopted centralized approaches \cite{centralized1,centralized2,centralized3}, in which a central node aggregates global information and plans trajectories for all agents. However, these methods face scalability limitations and impose high communication overhead, as all data must be processed by a single point. To overcome these issues, distributed MATP methods have gained attention \cite{DistributedSynch1,DistributedSynch2,DistributedSynch3}. These approaches distribute computation across agents to enhance scalability and reduce communication load. Nevertheless, most of them rely on synchronous updates and implicitly assume perfect communication, which may not hold in practice.

To tackle the potential imperfections in communication, asynchronous trajectory planning methods have been studied. \cite{kondo2023robust} and \cite{tordesillas2021mader} compute collision-avoidance motions in a decentralized and asynchronous manner using a check-recheck scheme based on a tight outer polyhedral representation constructed with the MINVO \cite{tordesillas2022minvo} basis. \cite{zhou2022swarm} proposes an MATP method that utilizes the GCOPTER \cite{wang2022geometrically} optimizer by incorportating collision constraints into the optimization cost. Although both methods can compute trajectories quickly, they do not guarantee mission success in complex environments, as potential deadlocks are not considered. Furthermore, inter-agent collision avoidance relies on access to the previously planned trajectories of other agents, which requires communication.
\cite{csenbacslar2024dream} introduces an asynchronous inter-robot collision avoidance strategy based on discretized separating hyperplane trajectories. The method involves minimal communication, requiring only the exchange of trajectory start times among agents. Although it ensures inter-agent collision avoidance, it does not address deadlock situations. \cite{zhang2025gcbf+} enables communication-free swarm coordination by leveraging graph control barrier functions under the assumption that each agent can obtain the positions of neighboring agents through onboard sensing. Similar to the aforementioned methods, it lacks guarantees for collision avoidance or deadlock resolution in obstacle-rich environments.

\begin{table}[t!]
\caption{Comparison with the State-of-the-Art Algorithms.}
\label{table:comparison_with_sota}
    \begin{center}
    \begin{tabularx}{\linewidth}{c|X|X|X|X}
    \hline
    \centering \multirow{2}{*}{Method} & \multicolumn{2}{c}{\textbf{Communication settings}}&\multicolumn{2}{c}{\textbf{Theoretical guarantee}} \tabularnewline
    & \centering Asynch. & \centering Com-free & \centering Colli-Av & \centering Dead-res \tabularnewline \hline
     MADER \cite{tordesillas2021mader} & \centering $\checkmark$ & \centering \xmark & \centering \xmark & \centering \xmark \tabularnewline
    EGO-v2 \cite{zhou2022swarm} & \centering $\checkmark$  & \centering \xmark & \centering \xmark & \centering \xmark  \tabularnewline
        DREAM \cite{csenbacslar2024dream} & \centering $\checkmark$ & \centering \xmark & \centering \xmark & \centering $\triangle$ \tabularnewline
    GCBF+ \cite{zhang2025gcbf+} & \centering $\checkmark$ & \centering $\checkmark$ & \centering \xmark & \centering \xmark \tabularnewline
    RBL \cite{boldrer2023rule} & \centering $\checkmark$  & \centering $\checkmark$ & \centering $\checkmark$ & \centering $\triangle$  \tabularnewline
    LTL-based \cite{alonso2018reactive} & \centering \xmark  & \centering \xmark & \centering  $\checkmark$ & \centering  $\checkmark$  \tabularnewline
    DLSC-GC \cite{ijrr_jungwon} & \centering \xmark & \centering \xmark & \centering $\checkmark$ & \centering $\checkmark$ \tabularnewline \hline
    \textbf{Ours$^{*}$} & \centering \cmarkbold & \centering \cmarkbold & \centering \cmarkbold & \centering \cmarkbold \tabularnewline
    \hline
    \end{tabularx}
    \end{center}
\footnotesize{
\{Asynch., Com-free, Colli-Av, Dead-res\} are short for \{Asynchronous, Communication-free, Collision avoidance, Deadlock resolution\}. \newline
    $\checkmark$ means that the algorithm explicitly considers the corresponding item.\newline
    $\triangle$ means that it considers the corresponding item, but in obstacle-free space.\newline
    $*$ means our no-communication-based method, MC-Swarm-N.
    } 
\end{table}

\subsection{Collision Avoidance and Deadlock Resolution in MATP}
\label{subsec2:deadlock_resolution}
In the field of MATP, collision avoidance is essential, and various methods for deadlock resolution have been studied to ensure goal convergence. 
\cite{deadlock_bvc1,deadlock_bvc2,deadlock_bvc3}, which avoid inter-agent collision using buffered Voronoi cells, attemp to resolve deadlocks by applying the right-hand rule.
\cite{luo2021grpavoid} proposes a group adaptation mechanism that divides the entire UAV group into subgroups for better collision avoidance and deadlock resolution, along with control strategies to prevent both intergroup and intragroup collisions.
\cite{mao2024multi} proposes an approach that implicitly resolves deadlocks by leveraging global path planning results.

Although the above methods work well in mild conditions (e.g., sparse-obstacle environment), there has been work aimed at theoretically guaranteeing their safe operation with a larger number of agents and their applicability in cluttered settings.
\cite{semnani2020force} based on force-based motion planning, guarantees that a safe distance is maintained between agents. However, in narrow spaces, this method can cuase agents to fail to move toward the goal and instead oscillate in place.
\cite{alonso2018reactive} employs a high-level mission planner that adopts the benefits of both linear temporal logic (LTL) and a local planner to resolve issues such as deadlock and collision. However, since the LTL-based module is fully centralized, a synchronized communication is required.
\cite{boldrer2023rule} is most similar to our work direction in that it does not require synchronization or communication between robots and provides theoretical guarantees for goal convergence and safety. It uses the Lloyd-based algorithm \cite{lloyd1982least} to address various issues, but it does not consider collision avoidance for obstacles or deadlock resolution taking obstacles into account. In contrast, we guarantee both safety and goal convergence in complex and narrow environments caused by obstacles.
\subsection{Comparison with the Previous Works}
\label{subsec2:comp_prev}
Over the past few years, we have proposed MATP methods focusing on theoretical guarantee of collision avoidance (CA) and dealock resolution (DR). The following are some of our representative works.
\begin{itemize}
    \item LSC \cite{LSC}: guarantees CA
    \item DLSC \cite{park2023dlsc}: guarantees CA
    \item DLSC-GC \cite{ijrr_jungwon}: guarantees CA and DR
\end{itemize}
Both LSC and DLSC perform CA based on linear safe corridors, but differ in their DR methods: LSC utilizes higher priority-based subgoal planning, while DLSC adopts mode-based subgoal planning. 
While DLSC shows superior goal convergence performance in narrow environments such as mazes, it lacks a theoretical guarantee of deadlock-freedom. In contrast, DLSC-GC guarantees safety using variants of safe corridors and deadlock resolution through linear programming-based subgoal planning, but it is designed under the assumption of perfect synchronization and communication. In this paper, we propose the MATP algorithm, which extends this approach to systems that operate under lightweight or even communication-free settings, while still ensuring CA and DR.
\section{Problem Formulation}
\label{sec3:problem_formulation}
Suppose that $N$ homogenous robots navigate in a static obstacle environment $\mathcal{O}$.
Our goal is to generate safe and deadlock-free trajectories while minimizing network communication during flight. 

Throughout this paper, we will use the notation in Table \ref{table: notation}. 
A calligraphic uppercase letter denotes a set (e.g., $\mathcal{I}$), a bold uppercase letter means a matrix (e.g., $\textbf{A}$), an italic letter means a scalar value (e.g., $r$), and a bold lowercase letter indicates a vector (e.g., $\textbf{x}$).
The superscript with parenthesis (e.g., $\textbf{x}^{(h)}$) indicates the state update step of the symbol, and the superscript will be omitted when the symbol is from the current state update step.
\subsection{Assumption}
\label{subsec: assumption}
In this paper, we make the following assumptions:
\begin{itemize}
    \item (Obstacle and sensing) The obstacle space $\mathcal{O}$ is provided as prior knowledge. Each agent can observe the positions of other agents and static obstacles without sensing delay. Systems equipped with ultra-wideband sensors \cite{zhou2022swarm} can be candidate options capable of tracking the positions of all agents.
    \item (Grid-based planner) All agents share the same grid space $G=(\mathcal{V},\mathcal{E})$, where the grid size $d$ is larger than $2\sqrt{2}r$ and $r$ is the radius of the agent. The agent positioned on the grid does not collide with static obstacles.
    \item (Mission) The agent's start point and desired goal are assigned to one of vertices of the grid space $G$, and different agents have different start point and desired goal.
    \item (Synchronization) 
    Each robot's clock is synchronized before navigation using a protocol such as Network Time Protocol (NTP) \cite{mills2010network}.
\end{itemize} 
\begin{table}[t!]
\caption{Nomenclature} 
\label{table: notation}
\begin{center}
\begin{tabularx}{\linewidth}{|s|d|}
\hline
\centering \textbf{Symbol} & \centering \textbf{Definition} \tabularnewline
\hline  
\centering $\textbf{x}^{(h)}$ & This superscript indicates that the symbol is sampled or planned at the state update step $h$. It is omitted if the symbol is from the current state update step. \tabularnewline
\hline
\centering $G=(\mathcal{V},\mathcal{E})$, $d$ & Grid space ($\mathcal{V}$: grid vertices, $\mathcal{E}$: grid edges), grid size $d>2\sqrt{2}r$ \tabularnewline
\hline
\centering $N$, $\mathcal{I}$ & The number of agents, set of agents $\mathcal{I} = \{1,\cdots,N\}$.\tabularnewline
\hline
\centering $\textbf{p}_{i}(t),\textbf{v}_{i}(t),\textbf{u}_{i}(t)$ &  Position, velocity, and
control input of agent $i$. \tabularnewline
\hline
\centering $v_{\max}, u_{\max}$ &  Agent's maximum velocity and acceleration. \tabularnewline
\hline
\centering $\mathcal{O}$ & Space occupied by the static obstacles. \tabularnewline
\hline
\centering $r$, $\mathcal{C}$ & Agent's radius, collision model. \tabularnewline
\hline
\centering $\textbf{z}_{i}, \textbf{s}_{i}$ & Desired goal and start point of agent $i$, respectively. \tabularnewline
\hline
\centering $h$, $T_{s}$, $t_{h}$ & Current state update step, state update period, and time at the state update step $h$. $t_{h} = t_{0} + h T_{s}$. \tabularnewline
\hline
\centering $T_{r}$ & Maximum trajectory replanning period. \tabularnewline
\hline
\centering $\hat{\textbf{p}}_{i}, \textbf{w}_{i}, \textbf{g}_{i}$ & Estimated position, waypoint, subgoal of agent $i$, respectively. \tabularnewline
\hline
\centering $\mathcal{S}_{i}$, $\mathcal{V}_{i}$ & Safe flight corridor (SFC), modified Buffered Voronoi Cell (BVC). \tabularnewline
\hline
\centering $M$, $\mathcal{M}$, $T$, $\Delta t$ & The number of steps to plan, $\mathcal{M} = \{0,\cdots,M\}$, planning horizon, and planning time step. Thus, $T = M \Delta t$. \tabularnewline
\hline
\centering $\|\textbf{x}\|$, $\|\textbf{x}\|_{\infty}$, $\lvert \boldsymbol{\pi}_{i} \rvert$ & Euclidean norm, L-infinity norm, makespan of the path $\boldsymbol{\pi}_{i}$. \tabularnewline
\hline
\centering $\textbf{0}$, $\oplus$, $\text{Conv}(\mathcal{X})$, $[a, b]$, $[\textbf{p}, \textbf{q}]$  & Zero vector, Minkowski sum, convex hull operator, closed scalar interval between $a$ and $b$, and line segment between two points $\textbf{p}$ and $\textbf{q}$. \tabularnewline
\hline
\end{tabularx}
\end{center}
\end{table}
\begin{figure*}[t!]
\centering
\includegraphics[width = 0.82\linewidth]{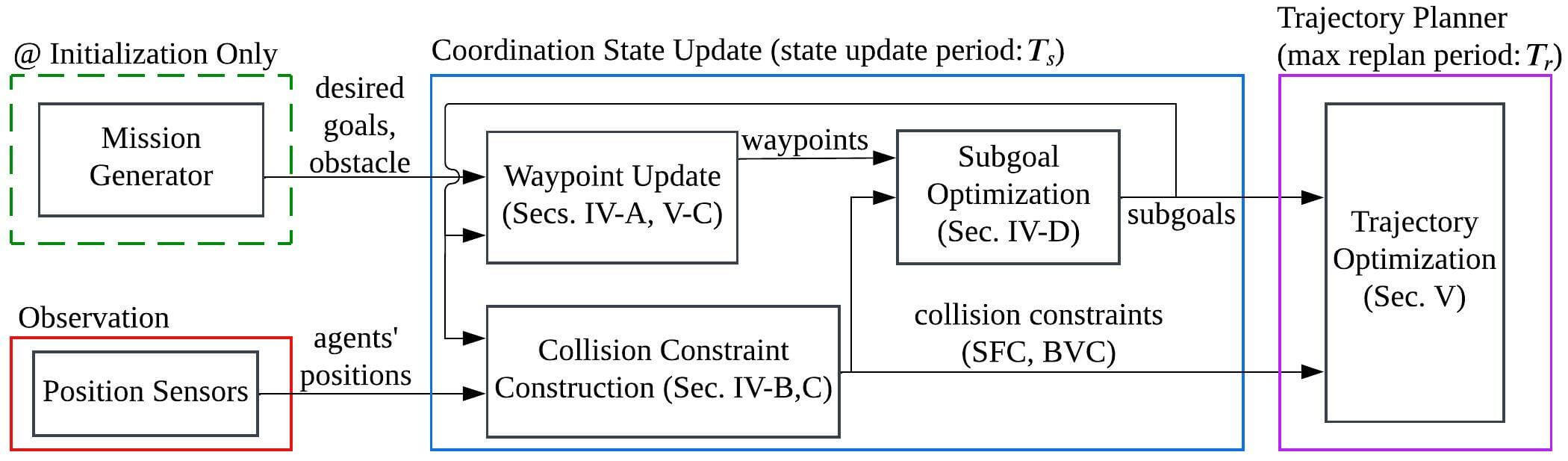}
\caption{Planning process of the proposed algorithm executed by each agent}
\label{fig: flow chart}
\vspace{-2mm}
\end{figure*}
\subsection{Agent Dynamics}
In this paper, we model agent $i \in \mathcal{I}$ in double-integrator dynamics as follows:
\begin{equation}
\label{eq: agent dynamics}
\begin{aligned} 
    \dot{\textbf{p}}_{i}(t) &= \textbf{v}_{i}(t) \\
    \dot{\textbf{v}}_{i}(t) &= \textbf{u}_{i}(t)
\end{aligned}
\end{equation}
where $\textbf{p}_{i}(t),\textbf{v}_{i}(t),\textbf{u}_{i}(t) \in \mathbb{R}^{3}$ are the position, velocity, and control input of agent $i$, respectively. The velocity and input constraints are given by:
\begin{equation}
\label{eq: velocity constraints}
    \|\textbf{v}_{i}(t)\|_{\infty} < v_{\max},\ \|\textbf{u}_{i}(t)\|_{\infty} < u_{\max}, \ \forall t
\end{equation}
where $\|\cdot\|_{\infty}$ is the L-infinity norm, and $v_{\max}, u_{\max}$ are the agent's maximum velocity and acceleration, respectively. 
\subsection{Collision Avoidance}
\subsubsection{Inter-agent collision avoidance}
The necessary condition for collision avoidance between the agents $i, j \in \mathcal{I}$ is given by:
\begin{equation}
     \|\textbf{p}_{i}(t) - \textbf{p}_{j}(t)\| \geq 2r, \quad \forall t
\label{eq: inter-agent collision avoidance}
\end{equation}
where $r$ is the radius of the agent.

\subsubsection{Static obstacle avoidance}
The necessary condition for collision avoidance between agent $i$ and static obstacles is given by:
\begin{equation}
    (\textbf{p}_{i}(t) \oplus \mathcal{C}) \cap \mathcal{O} = \emptyset, \forall t
\label{eq: static obstacle avoidance}
\end{equation}
\begin{equation}
  \mathcal{C}=\{\textbf{x} \in \mathbb{R}^{3} \mid \|\textbf{x}\| < r\}
\label{eq: static obstacle collision model}
\end{equation}
where $\oplus$ is the Minkowski sum, $\mathcal{O}$ is the space occupied by the static obstacles, and $\mathcal{C}$ is the sphere-shaped collision model.
\subsection{Overview}
\label{subsec3:overview}
Fig. \ref{fig: flow chart} illustrates the flowchart of the proposed method, which consists of two main modules. The first module is the coordination state update module, responsible for synchronizing the \textit{coordination state} among agents. In this work, the coordination state is defined as shared information necessary for collision avoidance and deadlock resolution, such as collision constraints, subgoals, and waypoints. 

To remove network communication during flight, each agent first plans waypoints using a grid-based multi-agent pathfinding (MAPF) algorithm (Sec. \ref{subsec: waypoint update}). Next, it generates a Safe Flight Corridor (SFC) and a modified Buffered Voronoi Cell (BVC) for collision avoidance (Secs. \ref{subsec: static obstacle avoidance} and \ref{subsec: inter-agent collision avoidance}). Finally, the subgoal is optimized using the waypoints and collision constraints to prevent deadlock (Sec. \ref{subsec: subgoal optimization}). 

The second module is the trajectory planning module, which allows each agent to asynchronously plan a safe and deadlock-free trajectory based on the updated coordination state (Sec. \ref{sec5:trajectory_planning}). These modules operate independently with different periods.
Specifically, the coordination state is updated at a fixed interval $T_{s}$, whereas the trajectory planning occurs at random intervals, up to a maximum replanning period of $T_{r}$.
The details of each module are described in the following sections.
\begin{algorithm}[t]
\SetAlgoLined
\KwIn{
Desired goal $\textbf{z}_{\forall i \in \mathcal{I}}$, start time $t_{0}$, and obstacle space $\mathcal{O}$}
\KwOut{Coordination state $\mathcal{H}$}
  waitUntil($t_{0}$)\;
  $h \gets 0$\;  
  \While{mission is not finished}{
    $\hat{\textbf{p}}_{\forall i \in \mathcal{I}} \gets$ detectAgentsPositions()\;
    $\textbf{w}_{\forall i \in \mathcal{I}} \gets$ updateWaypoints($\textbf{w}^{(h-1)}_{\forall i \in \mathcal{I}}$, $\textbf{g}^{(h-1)}_{\forall i \in \mathcal{I}}$, $\textbf{z}_{\forall i \in \mathcal{I}}$)\;
    \For{$\forall i \in \mathcal{I}$}{
      $\mathcal{S}_{i} \gets$ buildSFC($\hat{\textbf{p}}_{i}, \textbf{g}^{(h-1)}_{i}, \textbf{w}_{i}, \mathcal{O}$)\;
      $\mathcal{V}_{i} \gets$ buildBVC($\hat{\textbf{p}}_{\forall i \in \mathcal{I}}$, $\textbf{g}^{(h-1)}_{\forall i \in \mathcal{I}}$)\;
      $\textbf{g}_{i} \gets$ subgoalOpt($\textbf{g}^{(h-1)}_{i}$, $\textbf{w}_{i}$, $\mathcal{S}_{i}$, $\mathcal{V}_{i}$)\;
    }
    $\mathcal{H} \gets \{\textbf{w}_{\forall i \in \mathcal{I}}$, $\textbf{g}_{\forall i \in \mathcal{I}}$, $\mathcal{S}_{\forall i \in \mathcal{I}}$, $\mathcal{V}_{\forall i \in \mathcal{I}}\}$\;
    waitUntil($t_{h+1}$)\;
    $h \gets h + 1$\;
  }
\caption{Coordination state update}
\label{alg: coordination state update algorithm}
\end{algorithm}
\section{Coordination State Update}
\label{sec4:coordination_state_update}
In this paper, \textit{coordination state} is defined as the shared information among agents required for collision avoidance and deadlock resolution.
For instance, in MADER \cite{tordesillas2021mader} and DLSC \cite{park2023dlsc}, previously planned trajectories serve as the coordination state because they are used to generate collision constraints.
Synchronizing this coordination state is essential to ensure collision avoidance and deadlock resolution. 
However, many existing methods require extensive network communication to achieve this synchronization. 
To address this issue, we introduce a coordination state update method that does not rely on network communication during flight.

Algorithm \ref{alg: coordination state update algorithm} outlines the update process of the coordination state, which consists of waypoints, subgoals, and collision constraints. Each agent receives the agent's desired goals and the start time $t_{0}$ as input. After the start time $t_{0}$, each agent detects the current positions of all agents and updates the waypoints based on the previous waypoints, subgoals, and desired goals independently (lines 4-5 of Algorithm \ref{alg: coordination state update algorithm}, Sec. \ref{subsec: waypoint update}).
Next, collision constraints, such as Safe Flight Corridor (SFC) and modified Buffered Voronoi Cell (BVC), are constructed using only the current positions of the agents, which eliminates the need for inter-agent communication. (lines 7-8 of Algorithm \ref{alg: coordination state update algorithm}, Secs. \ref{subsec: static obstacle avoidance} and \ref{subsec: inter-agent collision avoidance}).
Subgoal optimization is then performed using the waypoints and collision constraints to prevent deadlocks (line 9 of Algorithm \ref{alg: coordination state update algorithm}, Sec. \ref{subsec: subgoal optimization}). 
The coordination state is updated with the new waypoints, subgoals, and collision constraints (line 11 of Algorithm \ref{alg: coordination state update algorithm}). 
This process is repeated at the state update period $T_{s}$ until the mission is completed. 

The proposed approach (MC-Swarm-N) requires network communication only during initialization, not during flight, to update the coordination state.
Therefore, it reduces network dependency compared to the previous works \cite{park2023decentralized, park2023dlsc}, which require continuous network communication to synchronize the coordination state. 

\subsection{Waypoint Update}
\label{subsec: waypoint update}
In this work, waypoints derived from a grid-based multi-agent pathfinding (MAPF) are used to guide agents toward their desired goals. 
Algorithm \ref{alg2:update_waypoints} presents the proposed waypoint update method.
First, each agent executes the MAPF algorithm on the grid space $G$ (line 1 of Algorithm \ref{alg2:update_waypoints}). 
The start points for the MAPF are assigned to the waypoints at the previous state update step $\textbf{w}^{(h-1)}_{i \in \mathcal{I}}$, and the goal points are set to the desired goals $\textbf{z}_{i \in \mathcal{I}}$. If it is the first time to run the MAPF, the start points are assigned to the agents' current position. 

In this work, we use Priority Inheritance with Backtracking (PIBT) \cite{okumura2022priority} for the MAPF algorithm because it is a fast and scalable algorithm that guarantees \textit{goal reachability}, meaning that the agent reaches its desired goal within a finite time. 
However, since PIBT is not an optimal MAPF algorithm, the makespan of the generated path may not decrease over time, potentially causing a livelock. To address this, we compare the makespan of the current path $\boldsymbol{\pi}_{\forall i \in \mathcal{I}}$ with the previous one $\boldsymbol{\pi}^{(h-1)}_{\forall i \in \mathcal{I}}$ and use the path with the shorter makespan (lines 2-4 of Algorithm \ref{alg2:update_waypoints}).
Next, we update the agent's waypoint to the second waypoint of the path (the waypoint one step after the start point) if the subgoals of all agents reach their respective waypoints (lines 5-6 of Algorithm \ref{alg2:update_waypoints}), i.e.:
\begin{equation}
\label{eq: waypoint update condition}
    \textbf{g}^{(h-1)}_{i} = \textbf{w}^{(h-1)}_{i}, \forall i\in \mathcal{I}
\end{equation}
where $\textbf{g}^{(h-1)}_{i}$ and $\textbf{w}^{(h-1)}_{i}$ are the subgoal and waypoint at the previous state update step, respectively.
If this condition is not met, we reuse the previous waypoint (lines 7-9 of Algorithm \ref{alg2:update_waypoints}).

After that, we check whether the agent has the same waypoint as other agents. If two agents have the same waypoint, one of them restores the waypoint to the previous one (lines 12-13 of Algorithm \ref{alg2:update_waypoints}). This process is repeated until there are no duplicated waypoints. 
Lemma \ref{lemma: no duplicated waypoints} presents that each agent has a different waypoint by the proposed algorithm.
\begin{lemma}
For the agents $i \in \mathcal{I}$ and $j \in \mathcal{I} \backslash \{i\}$, $\textbf{w}^{(h)}_{i} \neq \textbf{w}^{(h)}_{j}$ holds for every state update step $h$.
\label{lemma: no duplicated waypoints}
\end{lemma}
\begin{proof}
If $h=0$, all waypoints are the second waypoints of $\boldsymbol{\pi}_{\forall i \in \mathcal{I}}$ by line 6 in Algorithm \ref{alg2:update_waypoints}. Thus, the agent cannot have the same waypoint to the other agents.
Assume that there was no duplicated waypoints at the state update step $h-1$.
Then, agents $i$ and $j$ cannot have the same waypoints at the state update step $h$ because the agent's waypoint is restored to the previous one if duplicated waypoints are detected (See lines 10-17 in Algorithm \ref{alg2:update_waypoints}). Therefore, $\textbf{w}^{(h)}_{i} \neq \textbf{w}^{(h)}_{j}$ holds for every state update step $h$ by mathematical induction.
\end{proof}
\begin{algorithm}[t!]
\SetAlgoLined
\KwIn{Prev. waypoints $\textbf{w}^{(h-1)}_{\forall i \in \mathcal{I}}$, prev. subgoals $\textbf{g}^{(h-1)}_{\forall i \in \mathcal{I}}$, and desired goal $\textbf{z}_{\forall i \in \mathcal{I}}$}
\KwOut{Waypoint for current state update step $\textbf{w}_{\forall i \in \mathcal{I}}$}
  \tcp{Grid-based MAPF}
  $\boldsymbol{\pi}_{\forall i \in\mathcal{I}} \gets$ runMAPF($\textbf{w}^{(h-1)}_{\forall i \in \mathcal{I}}, \textbf{z}_{\forall i \in \mathcal{I}}$)\;
  \If{$h>0$ and $\lvert \boldsymbol{\pi}^{(h-1)}_{i} \rvert \leq \lvert \boldsymbol{\pi}_{i} \rvert$ \text{for} $\forall i\in\mathcal{I}$}{
      $\boldsymbol{\pi}_{\forall i\in\mathcal{I}} \gets \boldsymbol{\pi}^{(h-1)}_{\forall i \in \mathcal{I}}$
  }
  
  \tcp{Waypoint update}  
  \uIf{$h = 0$ or $\textbf{g}^{(h-1)}_{i} = \textbf{w}^{(h-1)}_{i}$ for $\forall i \in \mathcal{I}$}{
    $\textbf{w}_{\forall i \in \mathcal{I}} \gets$ second waypoint of $\boldsymbol{\pi}_{\forall i \in \mathcal{I}}$\;
  }
  \Else{
    $\textbf{w}_{\forall i \in \mathcal{I}} \gets \textbf{w}^{(h-1)}_{\forall i \in \mathcal{I}}$\;
  }

   \tcp{Conflict resolution}
   \If{$h>0$}{
     \For{$\forall i \in \mathcal{I}$}{
       \If{$\textbf{w}_{i} = \textbf{w}_{j}, \exists j \in \mathcal{I} 
           \backslash \{i\}$}{
         $\textbf{w}_{i} \gets \textbf{w}^{(h-1)}_{i}$\;
         \textbf{goto} line 10\;
       }
     }
   }

  \KwRet{$\textbf{w}_{\forall i \in \mathcal{I}}$}
\caption{Waypoints update}
\label{alg2:update_waypoints}
\end{algorithm}
\subsection{Static Obstacle Avoidance}
\label{subsec: static obstacle avoidance}
Safe flight corridor (SFC) \cite{flores2008real} is utilized to prevent collision with static obstacles. SFC is defined as a convex set that prevents the agent from a collision with static obstacles: 
\begin{equation}
  (\mathcal{S} \oplus \mathcal{C}) \cap \mathcal{O} = \emptyset
\label{eq: safe flight corridor definition}
\end{equation}
where $\mathcal{S}$ is the SFC, $\mathcal{C}$ is an obstacle collision model, $\mathcal{O}$ is the obstacle space, and $\oplus$ is the Minkowski sum.
In this work, we construct the SFC as follows:
\begin{equation}
\begin{alignedat}{2}
    \mathcal{S}_{i} = 
    \begin{cases} 
    \ \mathcal{S}(\{\hat{\textbf{p}}_{i}, \textbf{w}^{(0)}_{i}\})   & h = 0 \\
    \ \mathcal{S}(\{\hat{\textbf{p}}_{i}, \textbf{g}^{(h-1)}_{i}, \textbf{w}_{i}\})   & h > 0, (\ref{eq: safe flight corridor condition}) \\
    \ \mathcal{S}(\{\hat{\textbf{p}}_{i}, \textbf{g}^{(h-1)}_{i}\})   & h > 0, \text{else} \\
    \end{cases}
\end{alignedat}
\label{eq: safe flight corridor construction}
\end{equation}
\begin{equation}
    (\text{Conv}(\{\hat{\textbf{p}}_{i}, \textbf{g}^{(h-1)}_{i}, \textbf{w}_{i}\}) \oplus \mathcal{C}) \cap \mathcal{O} = \emptyset
\label{eq: safe flight corridor condition}
\end{equation}
where $\mathcal{S}_{i}$ is the SFC for agent $i$, $\mathcal{S}(\mathcal{P})$ is a convex set that includes the point set $\mathcal{P}$ and satisfies $(\mathcal{S}(\mathcal{P}) \oplus \mathcal{C}) \cap \mathcal{O} = \emptyset$, $\hat{\textbf{p}}_{i}$ is the estimated position of agent $i$ and $\text{Conv}(\cdot)$ is the convex hull operator that returns a convex hull of the input set.
In this work, $\mathcal{S}(\mathcal{P})$ is generated using the axis-search method \cite{park2020efficient}.
We first initialize the SFC as the axis-aligned bounding box that contains the point set $\mathcal{P}$. 
Then, we expand the SFC if the SFC is expandable in axis-aligned direction. This process is repeated until the SFC is not expandable.
Note that the SFC in (\ref{eq: safe flight corridor construction}) always contains the line segment between two points $\hat{\textbf{p}}_{i}$ and $\textbf{g}^{(h-1)}_{i}$. Therefore, the SFC does not block the agent when it tries to converge to the subgoal.

\subsection{Inter-Agent Collision Avoidance}
\label{subsec: inter-agent collision avoidance}
\begin{figure}[t!]
\centering
\includegraphics[width = 0.70\linewidth]{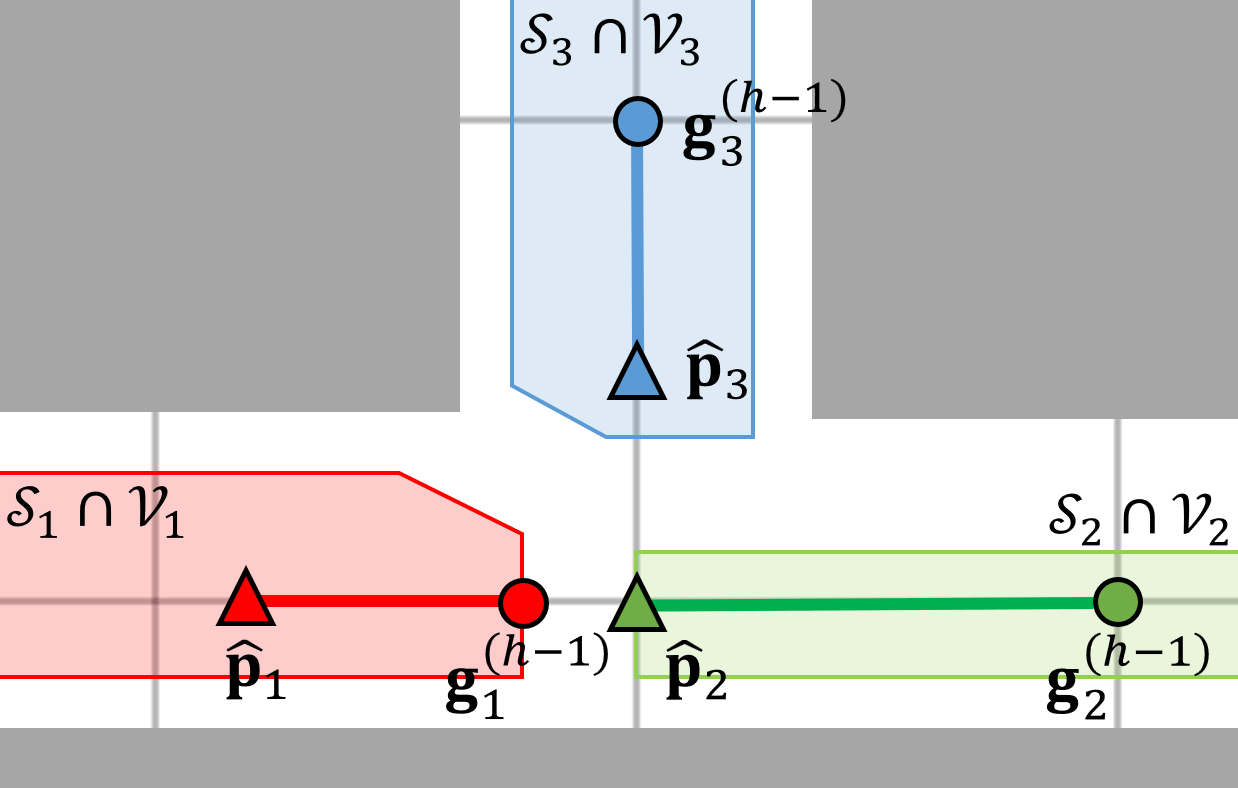}
\caption{
Collision constraints used in the proposed algorithm. The triangles are estimated positions, and the circles are subgoals at the previous state update step. The gray region represents static obstacles, and the color-shaded regions denote the feasible region that satisfies all collision constraints. 
}
\label{fig: collision constraints}
\end{figure}

Buffered Voronoi cell (BVC) \cite{zhou2017fast} is utilized for inter-agent collision avoidance. Original BVC is defined as a half-space that satisfies the following conditions:
\begin{subequations}
\begin{gather}
      \bar{\mathcal{V}}_{i} = \{\textbf{x} \in \mathbb{R}^{3} \mid (\textbf{x} - \hat{\textbf{p}}_{j}) \cdot \textbf{n}_{i,j} - d_{i,j} \geq 0, \forall j \neq i\} \\
      \textbf{n}_{i,j} = \frac{\hat{\textbf{p}}_{i}-\hat{\textbf{p}}_{j}}{\|\hat{\textbf{p}}_{i}-\hat{\textbf{p}}_{j}\|} \\
      d_{i,j} = r + \frac{1}{2}\|\hat{\textbf{p}}_{i}-\hat{\textbf{p}}_{j}\|
\end{gather}
\label{eq: bvc}%
\end{subequations}
where $\bar{\mathcal{V}}_{i}$ is the BVC, and $r$ is the agent's radius. 
In this paper, we modify the BVC, similar to \cite{park2023decentralized}, to enable the agents to converge to the subgoal more quickly:
\begin{subequations}
\begin{gather}
      \mathcal{V}_{i} = \{\textbf{x} \in \mathbb{R}^{3} \mid (\textbf{x} - \textbf{c}_{j,i}) \cdot \textbf{n}_{i,j} - d_{i,j} \geq 0, \forall j \neq i\} \label{eq: modified bvc1}\\
      \textbf{n}_{i,j} = \frac{\textbf{c}_{i,j}-\textbf{c}_{j,i}}{\|\textbf{c}_{i,j}-\textbf{c}_{j,i}\|} \label{eq: modified bvc2}\\
      d_{i,j} = r + \frac{1}{2}\|\textbf{c}_{i,j}-\textbf{c}_{j,i}\| \label{eq: modified bvc3}
\end{gather}
\label{eq: modified bvc}%
\end{subequations}
where $\mathcal{V}_{i}$ is the modified BVC, $\textbf{c}_{i,j}$ is the point on the line segment $[\hat{\textbf{p}}_{i}, \textbf{g}^{(h-1)}_{i}]$ that is closest to the line segment $[\hat{\textbf{p}}_{j}, \textbf{g}^{(h-1)}_{j}]$, and $[\textbf{a},\textbf{b}]$ is the line segment between two points $\textbf{a}$ and $\textbf{b}$. 
Similar to the SFC, modified BVC includes the line segment between $\hat{\textbf{p}}_{i}$ and $\textbf{g}^{(h-1)}_{i}$, as shown in Fig. \ref{fig: collision constraints}. Therefore, each agent can converge to its subgoal without being obstructed by collision constraints.
Lemma \ref{lemma: safety of bvc} shows that the modified BVC guarantees collision avoidance between the agents.
\begin{lemma}
If $\textbf{p}_{i} \in \mathcal{V}_{i}$, $\textbf{p}_{j} \in \mathcal{V}_{j}$, then $\|\textbf{p}_{i} - \textbf{p}_{j}\| \geq 2r$.
\label{lemma: safety of bvc}
\end{lemma}

\begin{proof}
Since $\textbf{p}_{i} \in \mathcal{V}_{i}$, $\textbf{p}_{j} \in \mathcal{V}_{j}$, we have the following inequalities by adding (\ref{eq: modified bvc1}) for the agents $i$, $j$:
\begin{equation}
\begin{aligned}
    (\textbf{p}_{i}-\textbf{c}_{j,i}) \cdot \textbf{n}_{i,j} 
    + (\textbf{p}_{j}-\textbf{c}_{i,j}) \cdot \textbf{n}_{j,i} \\
    - (d_{i,j} + d_{j,i}) \geq 0
\end{aligned}
\label{eq: bvc safety1}
\end{equation}
This can be simplified using $\textbf{n}_{i,j} = -\textbf{n}_{j,i}$, (\ref{eq: modified bvc2}), and (\ref{eq: modified bvc3}):
\begin{subequations}
    \begin{align}
        \label{eq: bvc safety2}
    \nonumber (\textbf{p}_{i}-\textbf{p}_{j}) \cdot \textbf{n}_{i,j} 
    + (\textbf{c}_{i,j} - \textbf{c}_{j,i}) \cdot \textbf{n}_{i,j} \\
    - (d_{i,j} + d_{j,i}) \geq 0,\\
        \label{eq: bvc safety3}
        (\textbf{p}_{i}-\textbf{p}_{j}) \cdot \textbf{n}_{i,j} \geq 2r
    \end{align}
\end{subequations}
Therefore, $\|\textbf{p}_{i} - \textbf{p}_{j}\| \geq (\textbf{p}_{i}-\textbf{p}_{j}) \cdot \textbf{n}_{i,j} \geq 2r$ holds by Cauchy–Schwarz inequality.
\end{proof}

\subsection{Subgoal Optimization}
\label{subsec: subgoal optimization}
A subgoal is an intermediate target point that allows agents to follow their waypoints without causing deadlock.
One simple way to determine a subgoal is to assign the waypoint as a subgoal directly. 
However, this approach may lead to deadlock because agents can be blocked by collision constraints before reaching the subgoal.
To address this issue, we introduce a subgoal optimization method that preemptively prevents deadlock.
First, we compute a feasible region for the agent that satisfies all collision constraints.
Next, we find the subgoal closest to the waypoint within this feasible region and on the grid. 
More precisely, the subgoal is determined by solving the following optimization problem:
\begin{equation}
\begin{aligned}
& \underset{\textbf{g}_{i}}{\text{minimize}}     & & \|\textbf{g}_{i} - \textbf{w}_{i}\| \\
& \text{subject to}   & & \textbf{g}_{i} \in [ \textbf{s}_{i},\textbf{w}_{i}] & & \text{if } h = 0 \\
&                     & & \textbf{g}_{i} \in [ \textbf{g}^{(h-1)}_{i},\textbf{w}_{i}] & & \text{if } h > 0  \\
&                     & & \textbf{g}_{i} \in \mathcal{S}_{i} \cap \mathcal{V}_{i}  \\
\end{aligned}
\label{eq: subgoal optimization}
\end{equation}
where $\textbf{g}_{i}$ is the subgoal, $\textbf{w}_{i}$ is the waypoint, $\textbf{s}_{i}$ is the start point of agent $i$, and $[\cdot,\cdot]$ represents a line segment connecting two points. 
Lemmas \ref{lemma: subgoal and waypoint on the same edge} and \ref{lemma: one agent per one edge} show the key properties of the proposed subgoal optimization.
Lemma \ref{lemma: subgoal and waypoint on the same edge} demonstrates that the subgoal and waypoint are always on the same grid edge, and Lemma \ref{lemma: one agent per one edge} indicates that the subgoals of two agents cannot be on the middle of the same grid edge. 
\begin{figure*}[t!]
    \centering
    \begin{subfigure}[t]{0.22\textwidth}
        \centering
        \includegraphics[width=\textwidth]{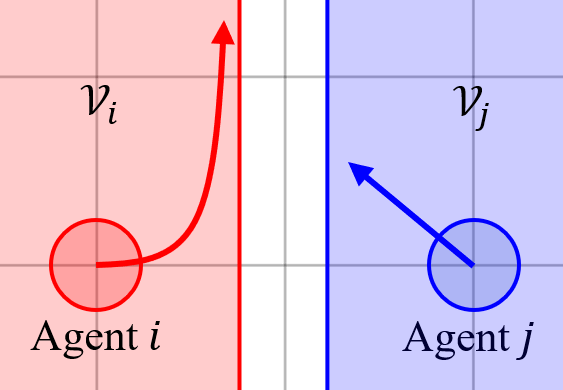}
        \caption{$t = t_{i}$}
        \label{fig: previous collision constraint1}
    \end{subfigure}
    \begin{subfigure}[t]{0.22\textwidth}
        \centering
        \includegraphics[width=\textwidth]{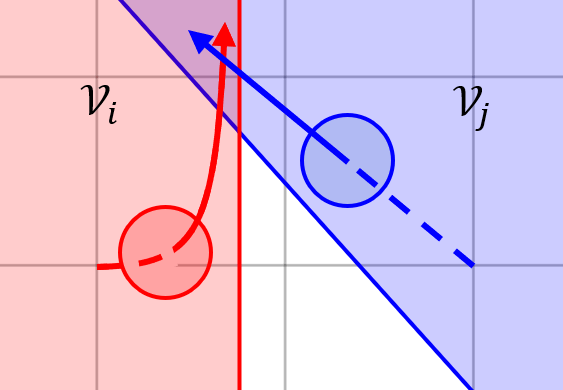}
        \caption{Naive method, $t = t_{j}$}
        \label{fig: previous collision constraint2}
    \end{subfigure}
    \begin{subfigure}[t]{0.22\textwidth}
        \centering
        \includegraphics[width=\textwidth]{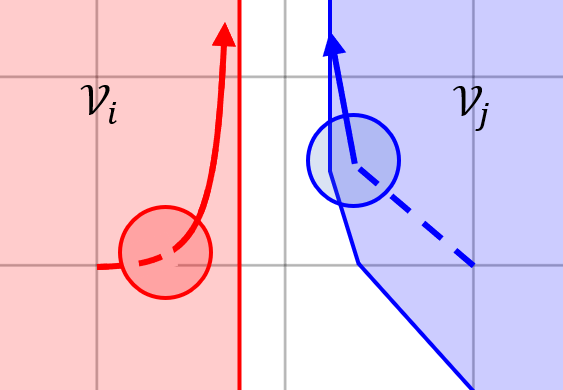}
        \caption{Proposed method, $t = t_{j}$}
        \label{fig: proposed collision constraint1}
    \end{subfigure}
    \caption{
     Comparison of trajectory planning when using the most recent collision constraint (naive method) and the conservative constraint (proposed method). The color-shared region represent collision constraint used for trajectory planning, and $t_{i}$ and $t_{j}$ are time when agents $i$ and $j$ update their trajectory, respectively.
     }
    \label{fig: collision constraint comparison}
    \vspace{-2mm}
\end{figure*}
\begin{lemma}
  $\textbf{g}_{i}$, $\textbf{g}^{(h-1)}_{i}$, and $\textbf{w}_{i}$ are always on the same grid edge $e \in \mathcal{E}$. 
\label{lemma: subgoal and waypoint on the same edge}
\end{lemma}

\begin{proof}
If $h = 0$, $\textbf{s}_{i}$ and $\textbf{w}_{i}(t_{0})$ are on the same grid edge because $\textbf{s}_{i}$ is on the grid vertex by the assumption and $\textbf{s}_{i}$ and $\textbf{w}_{i}(t_{0})$ are the consecutive waypoints of the path generated by the grid-based MAPF. Since $\textbf{g}_{i}(t_{0}) \in [\textbf{s}_{i}, \textbf{w}_{i}(t_{0})]$ holds by the constraint of (\ref{eq: subgoal optimization}), $\textbf{g}_{i}(t_{0})$ and $\textbf{w}_{i}(t_{0})$ are on the same grid edge.

Assume that $\textbf{g}^{(h-1)}_{i}$ and $\textbf{w}^{(h-1)}_{i}$ are on the same grid edge. 
If $\textbf{g}^{(h-1)}_{i} \neq \textbf{w}^{(h-1)}_{i}$, then $\textbf{w}_{i} = \textbf{w}^{(h-1)}_{i}$ holds by the waypoint update rule (\ref{eq: waypoint update condition}).
Thus, we obtain $[\textbf{g}_{i}, \textbf{w}_{i}] \subset [ \textbf{g}^{(h-1)}_{i}, \textbf{w}_{i} ] = [ \textbf{g}^{(h-1)}_{i}, \textbf{w}^{(h-1)}_{i} ]$ by the constraint in (\ref{eq: subgoal optimization}). It implies that $\textbf{g}_{i}$, $\textbf{g}^{(h-1)}_{i}$, and $\textbf{w}_{i}$ are on the same grid edge.
If $\textbf{g}^{(h-1)}_{i} = \textbf{w}^{(h-1)}_{i}$, then we have $[\textbf{g}_{i}, \textbf{w}_{i}] \subset [\textbf{g}^{(h-1)}_{i}, \textbf{w}_{i}] = [\textbf{w}^{(h-1)}_{i}, \textbf{w}_{i}]$. Since $\textbf{w}^{(h-1)}_{i}$ and $\textbf{w}_{i}$ are the consecutive waypoints of the path generated by the grid-based MAPF, $\textbf{g}_{i}$, $\textbf{g}^{(h-1)}_{i}$, and $\textbf{w}_{i}$ are on the same grid edge.
Thus, Lemma \ref{lemma: subgoal and waypoint on the same edge} holds by mathematical induction.
\end{proof}
\begin{lemma}
  If $\textbf{g}_{i}$ and $\textbf{g}_{j}$ are on the same grid edge $e \in \mathcal{E}$, then $\textbf{g}_{i}$ or $\textbf{g}_{j}$ must be on the vertex of the grid.
\label{lemma: one agent per one edge}
\end{lemma}

\begin{proof}
Assume that both $\textbf{g}_{i}$ and $\textbf{g}_{j}$ are not on the vertex of the grid. Then, there exists the replanning step $h_{i} \leq h_{j} < h$ that satisfies $\textbf{w}^{(h_{i})}_{i} \neq \textbf{w}^{(h_{i}+1)}_{i} = \cdots = \textbf{w}^{(h)}_{i}$ and $\textbf{w}^{(h_{j})}_{j} \neq \textbf{w}^{(h_{j}+1)}_{j} = \cdots = \textbf{w}^{(h)}_{j}$ without loss of generality. Here, we can observe that $\textbf{g}_{i} \in [\textbf{w}^{(h_{i})}_{i}, \textbf{w}^{(h_{i}+1)}_{i}]$ and $\textbf{g}_{j} \in [\textbf{w}^{(h_{j})}_{j}, \textbf{w}^{(h_{j}+1)}_{j}]$ due to the waypoint update rule (\ref{eq: waypoint update condition}).

If $h_{i} = h_{j}$, then we have $[\textbf{w}^{(h_{i})}_{i}, \textbf{w}^{(h_{i}+1)}_{i}] = [\textbf{w}^{(h_{i})}_{j}, \textbf{w}^{(h_{i}+1)}_{j}]$ because $\textbf{g}_{i}$ and $\textbf{g}_{j}$ are on the same grid edge, but not on the vertex of the grid. However, it is impossible because the path generated from MAPF algorithm does not cause collision.
If $h_{i} < h_{j}$, then $[\textbf{w}^{(h_{i})}_{i}, \textbf{w}^{(h_{i}+1)}_{i}] = [\textbf{w}^{(h_{j})}_{j}, \textbf{w}^{(h_{j}+1)}_{j}]$ holds because $\textbf{g}_{i}$ and $\textbf{g}_{j}$ are on the same grid edge, but not on the vertex of the grid. It implies that $\textbf{w}^{(h_{i}+1)}_{i} = \textbf{w}^{(h_{j})}_{i} = \textbf{w}^{(h_{j})}_{j}$ or $\textbf{w}^{(h_{i}+1)}_{i} = \textbf{w}^{(h_{j}+1)}_{i} = \textbf{w}^{(h_{j}+1)}_{j}$. However, both of them are impossible due to Lemma \ref{lemma: no duplicated waypoints}.
Therefore, $\textbf{g}_{i}$ or $\textbf{g}_{j}$ must be on the vertex of the grid.
\end{proof}

The proposed subgoal optimization method offers two main advantages. First, it ensures that each agent can reach its subgoal because the path to the subgoal cannot be blocked by collision constraints. Second, it prevents agents from blocking each other since it assigns the subgoal to a unique grid edge.
Additionally, the proposed subgoal optimization method does not cause optimization failure because $\textbf{g}^{(h-1)}_{i}$ satisfies all constraints of (\ref{eq: subgoal optimization}).

\section{Trajectory Planning}
\label{sec5:trajectory_planning}
\subsection{Trajectory Optimization}
Based on the coordination state, each agent plans its trajectory asynchronously. Even if agents share the same coordination states, it is challenging to guarantee collision avoidance without network communication because agents cannot know when other agents will plan their trajectories.
For instance, assume that agents $i$ and $j$ update their trajectories at time $t_{i}$ and $t_{j}$ ($t_{i}\leq t_{j}$), respectively . 
If the agents only use the most recent collision constraint when planning their trajectories, a collision may occur because there exists a region that both agents consider safe as shown in Fig. \ref{fig: previous collision constraint2}.

To address this issue, we introduce conservative collision constraints, inspired by discretized separating hyperplane trajectories (DSHT) \cite{csenbacslar2022asynchronous, csenbacslar2024dream}.
Suppose that the maximum trajectory replanning period is given by $T_{r}$.
To avoid collision, we plan the trajectory using the collision constraint generated from time $t-T_{r}-T_{s}$ to $t$, where $t$ is the current time and $T_{s}$ is state update period.
This ensures collision avoidance because the agents share at least one collision constraint, as shown in Fig. \ref{fig: proposed collision constraint1}.
This approach is similar to DSHT in using conservative collision constraints, but DSHT requires continuous network communication to update the constraints, whereas the proposed method does not.

For trajectory optimization, we formulate the cost functions to minimize the distance to the subgoal and the acceleration of the trajectory.
The trajectory must satisfy the input and velocity constraints for dynamical feasibility, and the agent should stop at the end of planning horizon to prevent collision in case of optimization failure.
To summarize, the trajectory optimization problem for agent $i$ at time $t$ is formulated as the following quadratic programming (QP) problem:
\begin{equation}
\begin{aligned}
& \underset{\textbf{u}_{i,k}}{\text{min}} & & J_{i}(\textbf{u}_{i,k}) & & \\
& \text{s.t.}   & & \textbf{x}_{i,k+1} = \textbf{A}\textbf{x}_{i,k} + \textbf{B}\textbf{u}_{i,k} & & \forall k \in \{0,\cdots,M-1\}\\ 
&                     & & \textbf{p}_{i,k} \in \mathcal{S}^{(h)}_{i} \cap \mathcal{V}^{(h)}_{i}  & & \forall k \in \mathcal{M}, \forall h \text{ s.t.}\\
&                     & & & &t_{h} \in [t-T_{r}-T_{s}, t] \\
&                     & & \|\textbf{v}_{i,k}\|_{\infty} < v_{max} & & \forall k \in \mathcal{M} \\
&                     & & \|\textbf{u}_{i,k}\|_{\infty} < u_{max} & & \forall k \in \mathcal{M} \\
&                     & & \textbf{v}_{i,M} = \textbf{0} & & 
\end{aligned}
\label{eq: trajectory optimization}
\end{equation}
\begin{equation}
J_{i}(\textbf{u}_{i,k}) = w_{e} \|\textbf{p}_{i,M}-\textbf{g}^{(h_{l})}_{i}\|^{2} + w_{a} \sum\limits_{k=0}^{M-1} \|\textbf{u}_{i,k}\|^{2}
\label{eq: cost function}
\end{equation}
where $J_{i}$ is the cost function, $\textbf{x}_{i,k} = [\textbf{p}_{i,k}^{T}, \textbf{v}_{i,k}^{T}]^{T}$, $\textbf{p}_{i,k}$, $\textbf{v}_{i,k}$, and $\textbf{u}_{i,k}$ are the state, position, velocity, and control input of agent $i$ at time step $k$, respectively. $M$ is the number of steps to plan, $\mathcal{M} = \{0,\cdots,M\}$, $T_{r}$ is the maximum replanning period, $T_{s}$ is the state update period, and $\mathcal{S}^{(h)}_{i}$ and $\mathcal{V}^{(h)}_{i}$ are the SFC and the modified BVC at the state update step $h$, respectively. $\textbf{A}$ and $\textbf{B}$ are matrices that represents the agent dynamics (\ref{eq: agent dynamics}), $\textbf{0} \in \mathbb{R}^{3}$ is the zero vector, $v_{\max}$ and $u_{\max}$ are the agent's maximum velocity and acceleration, respectively. $h_{l}$ is the latest state update step, and $w_{e}, w_{a} > 0$ are the weight coefficients.
This problem can be solved by using a conventional convex solver. 

Similar to original BVC \cite{zhou2017fast}, the proposed algorithm may cause optimization failure due to infeasible constraints. In such cases, the agent follows the previously planned trajectory. We note that the proposed algorithm ensures collision avoidance regardless of optimization failure, thanks to the final stop constraint ($\textbf{v}_{i,M} = \textbf{0}$) (Please refer to Theorem \ref{theorem: collision avoidance with optimization failure}).

\subsection{Theoretical Guarantee}
\label{subsec: theoretical guarantee}
This section describes the theoretical guarantee of the proposed method.
We will denote the planning horizon of trajectory optimization problem (\ref{eq: trajectory optimization}) by $T$ and the planning time step by $\Delta t$. Hence, $T = M \Delta t$ holds.

\subsubsection{Collision avoidance}
\label{subsubsec: collision avoidance}
Theorem \ref{theorem: collision avoidance} shows the sufficient condition for collision avoidance of the proposed algorithm.
\begin{theorem}
If the trajectory replanning period is less than the maximum replanning period $T_{r}$ and there is no optimization failure, then the agents do not collide with other agents or static obstacles.
\label{theorem: collision avoidance}
\end{theorem}

\begin{proof}
(Static obstacle avoidance)
There exists the state update step $h^{*}$ that satisfies $t_{h} \in [t-T_{r}-T_{s}, t]$ because the length of the interval is equal to or greater than the state update period $T_{s}$. Therefore, 
$(\textbf{p}_{i,k} \oplus \mathcal{C}) \cap \mathcal{O} \subset (\mathcal{S}^{(h^{*})}_{i} \oplus \mathcal{C}) \cap \mathcal{O} = \emptyset$ holds for $\forall i \in \mathcal{I}, k \in \mathcal{M}$ due to the constraint of (\ref{eq: trajectory optimization}). Thus, all agents do not collide with static obstacles by (\ref{eq: static obstacle avoidance}).

(Inter-agent collision avoidance)
Suppose that the latest trajectory planning time of the agents $i$ and $j$ are $t_{i}$ and $t_{j}$, respectively, and $t_{i} \leq t_{j}$ without loss of generality.
Then, $t_{j} - t_{i} < T_{r}$ holds because the replanning period is less than $T_{r}$.
Assume that the state update step $h^{*}$ satisfies $t_{h^{*}} \in [t_{i} - T_{s}, t_{i}]$. Note that such state update step must exist since the length of the interval $[t_{i} - T_{s}, t_{i}]$ is equal to the state update period $T_{s}$. Then, we obtain $t_{h^{*}} \in [t_{i} - T_{s}, t_{i}] \subset [t_{j} - T_{r} - T_{s}, t_{i}] = [t_{i} - T_{r} - T_{s}, t_{i}] \cap [t_{j} - T_{r} - T_{s}, t_{j}]$ because $t_{j} - t_{i} < T_{r}$.
This means that $\textbf{p}_{i,k} \in \mathcal{V}^{(h^{*})}_{i}$ and $\textbf{p}_{j,k} \in \mathcal{V}^{(h^{*})}_{j}$ holds for $\forall k \in \mathcal{M}$ due to the constraint of (\ref{eq: trajectory optimization}).
Thus, there is no collision between agents by Lemma \ref{lemma: safety of bvc}.
\end{proof}

Moreover, the proposed algorithm guarantees collision avoidance regardless of the optimization failure if the planning horizon is less than the maximum replanning period.
\begin{theorem}
If the planning horizon $T$ and trajectory replanning period are less than the maximum replanning period $T_{r}$, then the agents do not cause collision regardless of the optimization failure.
\label{theorem: collision avoidance with optimization failure}
\end{theorem}

\begin{proof}
Despite of optimization failure, the agents do not collide with static obstacles because the previously planned trajectories do not collide with static obstacles.
Suppose that the latest trajectory planning time of the agents $i$ and $j$ are $t_{i}$ and $t_{j}$, respectively, and $t_{i} \leq t_{j}$ without loss of generality.

(Case 1) If $t_{j} - t_{i} < T$, then there exists the state update step $h^{*}$ that satisfies $t_{h^{*}} \in [t_{i} - T_{s}, t_{i}]$. Therefore, we obtain $t_{h^{*}} \in [t_{i} - T_{s}, t_{i}] \subset [t_{j} - T_{r} - T_{s}, t_{i}] = [t_{i} - T_{r} - T_{s}, t_{i}] \cap [t_{j} - T_{r} - T_{s}, t_{j}]$ because $t_{j} - t_{i} < T < T_{r}$.
This means that $\textbf{p}_{i,k} \in \mathcal{V}^{(h^{*})}_{i}$ and $\textbf{p}_{j,k} \in \mathcal{V}^{(h^{*})}_{j}$ holds for $\forall k \in \mathcal{M}$ due to the constraint of (\ref{eq: trajectory optimization}).
Thus, there is no collision between agents by Lemma \ref{lemma: safety of bvc}.

(Case 2) If $t_{j} - t_{i} \geq T$, then agent $i$ stops at $\textbf{p}_{i,M}$ due to the constraint ($\textbf{v}_{i,M} = \textbf{0}$) in (\ref{eq: trajectory optimization}). Here, we can observe that $\textbf{p}_{i,M} \in \mathcal{V}^{(h)}_{i}$ holds for $\forall h$ such that $t_{h} > t_{i} + T$ because $\textbf{p}_{i,M} = \hat{\textbf{p}}^{(h)}_{i} \in \mathcal{V}^{(h)}_{i}$ if $t_{h} > t_{i} + T$. This implies that there exists the state update step $h^{*}$ that satisfies $\textbf{p}_{i,M} \in \mathcal{V}^{(h^{*})}_{i}$ and $\textbf{p}_{j,k} \in \mathcal{V}^{(h^{*})}_{j}$ for $\forall k \in \mathcal{M}$.
Thus, there is no collision between agents by Lemma \ref{lemma: safety of bvc}.
\end{proof}

\subsubsection{Deadlock resolution}
\label{subsubsec: deadlock resolution}
We define agent $i$ to be in a deadlock if it remains stationary at its current position and does not reach its target, i.e.:
\begin{equation}
    \textbf{p}_{i}(t) = \textbf{p}_{i}(t_{d}) \neq \textbf{z}_{i}, \forall t > t_{d} 
\label{eq: definition of deadlock}
\end{equation}
where $t_{d}$ is the time when the deadlock starts.
Lemmas \ref{lemma: subgoal convergence} and \ref{lemma: sufficient condition for deadlock resolution} present the sufficient condition for subgoal convergence and deadlock resolution of the proposed algorithm, respectively. 

\begin{lemma}
If the mission is solvable for the grid-based MAPF, there exists a state update step $h_{d}$ such that $\textbf{g}_{i}$ converges to $\textbf{g}^{(h_{d})}_{i}$ for $\forall i \in \mathcal{I}$.
\label{lemma: subgoal convergence}
\end{lemma}

\begin{proof}
Assume that there does not exist a state update step $h_{d}$ such that $\textbf{g}^{(h)}_{i}$ converges to $\textbf{g}^{(h_{d})}_{i}$.
Then, $\textbf{g}_{i}$ converges to $\textbf{w}_{i}$ because the cost function of the subgoal optimization (\ref{eq: subgoal optimization}), $\|\textbf{g}_{i} - \textbf{w}_{i}\|$, is strictly decreasing over time due to the constraint $\textbf{g}_{i} \in [\textbf{g}^{(h-1)}_{i}, \textbf{w}_{i}]$.
Moreover, $\textbf{w}_{i}$ converges to a specific point because we only use the path $\boldsymbol{\pi}_{i}$ whose makespan is monotonically decreasing when we update the waypoint (See lines 2-4 of Algorithm \ref{alg2:update_waypoints}).
Therefore, $\textbf{g}_{i}$ converges to a specific point, which contradicts the assumption.
Thus, there must exists a state update step $h_{d}$ such that $\textbf{g}_{i}$ converges to $\textbf{g}^{(h_{d})}_{i}$ for $\forall i \in \mathcal{I}$.
\end{proof}

\begin{lemma}
If there exists agent $i$ that satisfies $\hat{\textbf{p}}_{i} \neq \textbf{g}_{i}$, then the proposed algorithm does not cause deadlock. 
\label{lemma: sufficient condition for deadlock resolution}
\end{lemma}

\begin{proof}
If deadlock occurs, then the position and subgoal of the agents are converged to specific points by the definition of deadlock and Lemma \ref{lemma: subgoal convergence}.
Assume that $\hat{\textbf{p}}_{i}$ and $\textbf{g}_{i}$ converge to $\textbf{p}_{i,d}$ and $\textbf{g}_{i,d}$ after time $t_{d}$, respectively, and $\textbf{p}_{i,d} \neq \textbf{g}_{i,d}$.
Then, the optimal solution of (\ref{eq: trajectory optimization}), $\textbf{u}^{*}_{i,k}$, must satisfy $\textbf{u}^{*}_{i,k} = \textbf{0}$ for $\forall k \in \mathcal{M}$ after time $t_{d}$.

Let us define the following control input $\tilde{\textbf{u}}_{i,k}$:
\begin{equation}
    \begin{alignedat}{2}
    \tilde{\textbf{u}}_{i,k} = 
    \begin{cases} 
    \ \lambda \textbf{n}_{i,d} & k = 0 \\
    \ -\lambda \textbf{n}_{i,d}   &  k = 1 \\  
    \ \textbf{0}   &  k > 1 \\  
    \end{cases}
\end{alignedat}
\label{eq: convergence to the subgoal1}
\end{equation}
\begin{equation}
\textbf{n}_{i,d} = \frac{\textbf{g}_{i,d}-\textbf{p}_{i,d}}{\|\textbf{g}_{i,d}-\textbf{p}_{i,d}\|}
\label{eq: convergence to the subgoal2}
\end{equation}
where $\lambda > 0$. We can prove that $\tilde{\textbf{u}}_{i,k}$ is one of feasible solutions of the trajectory optimization problem (\ref{eq: trajectory optimization}) if $\lambda$ is small enough.
First, $[\textbf{p}_{i,d}, \textbf{g}_{i,d}] = [\hat{\textbf{p}}^{(h)}_{i}, \textbf{g}^{(h)}_{i}] \in \mathcal{S}^{(h)}_{i} \cap \mathcal{V}^{(h)}_{i}$ holds for all $h$ such that $t_{h} > t_{d}$. Therefore, $\tilde{\textbf{u}}_{i,k}$ satisfies the collision avoidance constraint since the trajectory is always on the line segment $[\textbf{p}_{i,d}, \textbf{g}_{i,d}]$.
Second, $\tilde{\textbf{u}}_{i,k}$ satisfies the velocity and input constraints if $\lambda < u_{\max}$ and $\lambda \Delta t < v_{\max}$. 
Third, $\tilde{\textbf{u}}_{i,k}$ satisfies the final stop constraint since the agent follows double-integrator dynamics.
Thus, $\tilde{\textbf{u}}_{i,k}$ is a feasible solution of (\ref{eq: trajectory optimization}) if $\lambda < u_{\max}$ and $\lambda < v_{\max}/\Delta t$.

The cost difference between $\textbf{u}^{*}_{i,k}$ and $\tilde{\textbf{u}}_{i,k}$ is given by: 
\begin{equation}
\begin{aligned}
    & J_{i}(\textbf{u}^{*}_{i,k}) - J_{i}(\tilde{\textbf{u}}_{i,k}) \\ 
    &= w_{e}d_{i}^{2} - w_{e}(d_{i} - \lambda \Delta t^{2})^{2} - w_{a}(2\lambda^{2}) \\ 
    &= w_{e} (2d_{i} \lambda \Delta t^{2} - \lambda^{2} \Delta t^{4}) - w_{a} (2\lambda^{2})\\
    &= (2w_{e}d_{i}\Delta t^{2}) \lambda - (w_{e} \Delta t^{4} + 2w_{a}) \lambda^{2} \\
    &= a \lambda - b \lambda^{2} 
\label{eq: deadlock sufficient condition8-1}
\end{aligned}
\end{equation}
where $d_{i} = \|\textbf{g}_{i,d}-\textbf{p}_{i,d}\|$, $a = 2w_{e}d_{i}\Delta t^{2} > 0$, and $b = w_{e} \Delta t^{4} + 2w_{a} > 0$. If $0 < \lambda < a/b$, we obtain $J_{i}(\textbf{u}^{*}_{i,k}) - J_{i}(\tilde{\textbf{u}}_{i,k}) > 0$. 
Therefore, $\textbf{u}^{*}_{i,k} = \textbf{0}$ is not the optimal solution, which contradicts the assumption that $\hat{\textbf{p}}_{i}$ converges to $\textbf{p}_{i,d}$ where $\textbf{p}_{i,d} \neq \textbf{g}_{i,d}$.
Therefore, the proposed algorithm does not cause deadlock if there exists agent $i$ that satisfies $\hat{\textbf{p}}_{i} \neq \textbf{g}_{i}$.
\end{proof}

Using Lemmas \ref{lemma: subgoal convergence} and \ref{lemma: sufficient condition for deadlock resolution}, we can prove that the proposed algorithm guarantees deadlock resolution.
\begin{theorem}
If the mission is solvable for the grid-based MAPF and $d > 2\sqrt{2}r$, then the proposed algorithm does not cause deadlock.
\label{theorem: deadlock resolution}
\end{theorem}
\begin{proof}
Assume that the proposed algorithm causes deadlock after time $t_{d}$. Then, the agents satisfy the following condition by Lemmas \ref{lemma: subgoal convergence} and \ref{lemma: sufficient condition for deadlock resolution}:
\begin{equation}
  \hat{\textbf{p}}^{(h)}_{i} = \textbf{g}^{(h)}_{i} = \textbf{g}^{(h_{d})}_{i} \qquad  \forall i \in \mathcal{I}, \forall h > h_{d}
\label{eq: necessary condition for deadlock}
\end{equation}
where $h_{d}$ is the state update step when deadlock happens.
The subgoal optimization problem (\ref{eq: subgoal optimization}) of agent $i \in \mathcal{I}$ can be reformulated as follows:
\begin{equation}
\begin{aligned}
& \text{minimize}     & & \delta \\
& \text{subject to}   & & \delta \in [0,1] \\
&                     & & \textbf{w}_{i} + \delta (\textbf{g}^{(h-1)}_{i} - \textbf{w}_{i}) \in \mathcal{S}_{i} \cap \mathcal{V}_{i} 
\end{aligned}
\label{eq: deadlock resolution1}
\end{equation}
where $\delta$ is a variable such that $\textbf{g}_{i} = \textbf{w}_{i} + \delta (\textbf{g}^{(h-1)}_{i} - \textbf{w}_{i})$. 
By (\ref{eq: necessary condition for deadlock}), we have $\text{Conv}(\{\hat{\textbf{p}}_{i}, \textbf{g}^{(h-1)}_{i}, \textbf{w}_{i}\}) = [\textbf{g}^{(h-1)}_{i}, \textbf{w}_{i}]$. This implies that the condition (\ref{eq: safe flight corridor condition}) is satisfied because $\textbf{g}^{(h-1)}_{i}$ and $\textbf{w}_{i}$ are always on the same grid edge by Lemma \ref{lemma: subgoal and waypoint on the same edge} and this grid edge does not collide with static obstacles. 
Therefore, the SFC constraint in (\ref{eq: deadlock resolution1}) can be omitted since $\textbf{g}_{i} \in [\textbf{g}^{(h-1)}_{i}, \textbf{w}_{i}] \subset \mathcal{S}_{i}$.
\begin{figure}[t!]
    \centering
    \includegraphics[width = 0.6\linewidth]{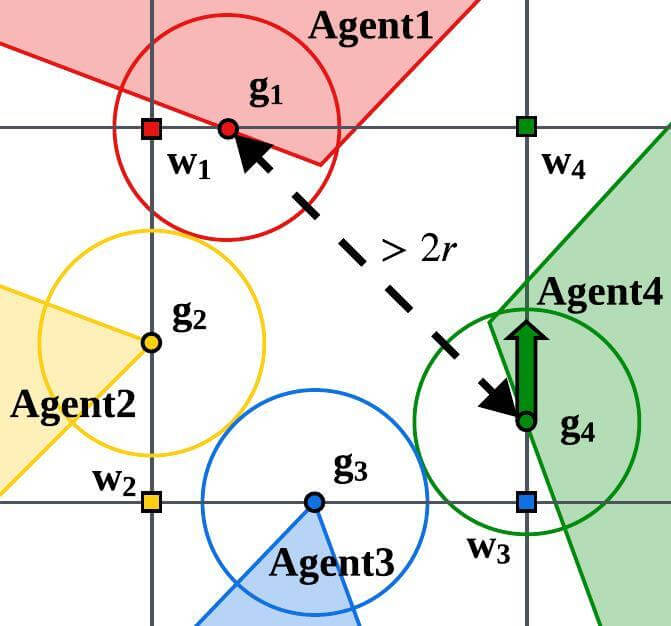}
    \caption{Example of blocking agents. The square dots, circle dots, and color-shaded regions represent waypoints, subgoals, and agents' feasible regions, respectively.
    }
    \label{fig: proof1}
\end{figure}
Accordingly, the Lagrangian function of (\ref{eq: deadlock resolution1}) for agent $i \in \mathcal{I}$ is formulated as follows:
\begin{equation}
\begin{aligned}
L &= \delta - \lambda_{0} \delta + \lambda_{1} (\delta - 1) \\
  &+ \sum_{j \in \mathcal{I}\backslash\{i\}} \lambda_{i,j} (d_{i,j} - (\textbf{w}_{i} + \delta (\textbf{g}^{(h-1)}_{i} - \textbf{w}_{i}) - \textbf{c}_{j,i}) \cdot \textbf{n}_{i,j})
\end{aligned}
\label{eq: deadlock resolution2}
\end{equation}
where $\lambda_{0}$, $\lambda_{1}$, and $\lambda_{i,j}$ are the Lagrangian multipliers.
The stationary condition of KKT conditions \cite{boyd2004convex} is given by:
\begin{equation}
\frac{\partial L}{\partial \delta} = 1 - \lambda_{0} + \lambda_{1} - \sum_{j \in \mathcal{I}\backslash\{i\}} \lambda_{i,j} (\textbf{g}^{(h-1)}_{i} - \textbf{w}_{i}) \cdot \textbf{n}_{i,j} = 0
\label{eq: deadlock resolution4}
\end{equation}
If $h>h_{d}$, then $\delta^{*} = 1$ is the optimal solution of the subgoal optimization problem (\ref{eq: deadlock resolution1}) due to (\ref{eq: necessary condition for deadlock}). Therefore, $\lambda_{0} = 0$ holds by the complementary slackness condition of KKT conditions.
Moreover, we have $\textbf{g}^{(h-1)}_{i} = \textbf{g}_{i}$, $\textbf{c}_{i,j} = \textbf{g}_{i}$, and $\textbf{c}_{j,i} = \textbf{g}_{j}$ due to (\ref{eq: necessary condition for deadlock}).
Thus, we can simplify (\ref{eq: deadlock resolution4}) as follows:
\begin{equation}
\sum_{j \in \mathcal{I}\backslash\{i\}} \lambda_{i,j} \frac{(\textbf{g}_{j} - \textbf{g}_{i})^{T}(\textbf{w}_{i} - \textbf{g}_{i})}{\|\textbf{g}_{i} - \textbf{g}_{j}\|} = 1 + \lambda_{1}
\label{eq: deadlock resolution5}
\end{equation}
To fulfill the above condition, there must exist an agent $j \in \mathcal{I}\backslash\{i\}$ that satisfies $\lambda_{i,j} > 0$ and $(\textbf{g}_{j} - \textbf{g}_{i})^{T}(\textbf{w}_{i} - \textbf{g}_{i}) > 0$ because $\lambda_{1} \geq 0$ and $\lambda_{i,j} \geq 0$ by the dual feasibility of KKT conditions.
Furthermore, $\lambda_{i,j} > 0$ implies that the following equations hold due to the complementary slackness of KKT conditions:
\begin{equation}
  d_{i,j} - (\textbf{w}_{i} + (\textbf{g}^{(h-1)}_{i} - \textbf{w}_{i}) - \textbf{c}_{j,i}) \cdot \textbf{n}_{i,j} = 0
\label{eq: deadlock resolution7}
\end{equation}
\begin{equation}
  r + \frac{1}{2}\|\textbf{g}_{i}-\textbf{g}_{j}\| - (\textbf{w}_{i} + (\textbf{g}_{i} - \textbf{w}_{i}) - \textbf{g}_{j}) \cdot \frac{\textbf{g}_{i} - \textbf{g}_{j}}{\|\textbf{g}_{i} - \textbf{g}_{j}\|} = 0
\label{eq: deadlock resolution8}
\end{equation}
\begin{equation}
  \|\textbf{g}_{i} - \textbf{g}_{j}\| = 2r
\label{eq: deadlock resolution9}
\end{equation}

To summarize, if the proposed algorithm causes deadlock after time $t_{d}$, there must exist an agent $j$ that satisfies the following conditions for each agent $i \in \mathcal{I}$:
\begin{equation}
  (\textbf{g}_{j} - \textbf{g}_{i})^{T}(\textbf{w}_{i} - \textbf{g}_{i}) > 0
\label{eq: deadlock resolution9-1}
\end{equation}
\begin{equation}
  \|\textbf{g}_{i} - \textbf{g}_{j}\| = 2r
\label{eq: deadlock resolution9-2}
\end{equation}

Let us define agent $B(i) \in \mathcal{I}$ that satisfies (\ref{eq: deadlock resolution9-1}) and (\ref{eq: deadlock resolution9-2}) as a \textit{blocking agent} of agent $i$, where $B(\cdot)$ indicates the blocking agent of the given agent. 
Fig. \ref{fig: proof1} illustrates the example of the blocking agent. In the figure, the yellow agent is the blocking agent of the red agent because the yellow agent prevents the red agent from reaching the waypoint $\textbf{w}_{1}$. Similarly, the green agent is the blocking agent of the yellow agent, and the blue agent is the blocking agent of the green agent.

Since there are a finite number of agents in $\mathcal{I}$, there must exist an agent $q \in \mathcal{I}$ that satisfies the following:
\begin{equation}
  \Delta(q) \geq \Delta(i), \forall i \in \mathcal{I}
\label{eq: deadlock resolution10}
\end{equation}
where $\Delta(i) = \|\textbf{w}_{i} - \textbf{g}_{i}\|$.
As discussed earlier, agent $q$ must have its blocking agent $B(q) \in \mathcal{I}$, and $\textbf{g}_{q}$ and $\textbf{g}_{B(q)}$ must be on different grid edges by Lemma \ref{lemma: one agent per one edge}. Therefore, $\Delta(B(q))$ is given by:
\begin{equation}
\begin{alignedat}{2}
    \Delta(B(q)) = 
    \begin{cases} 
    \ d - 2r + \Delta(q), &\text{if } D > 0  \\
    \ d - \sqrt{4r^{2} - \Delta(q)^{2}}, &\text{else} \\
    \end{cases}
\end{alignedat}
\label{eq: deadlock resolution11}
\end{equation}
where $D$ is defined as follows:
\begin{equation}
  D = (\textbf{w}_{B(q)} - \textbf{g}_{B(q)})^{T}(\textbf{w}_{q} - \textbf{g}_{q})
\label{eq: deadlock resolution12}
\end{equation}
If $d > 2\sqrt{2}r$, we obtain $\Delta(B(q)) > \Delta(q)$, which contradicts the assumption that agent $q$ satisfies (\ref{eq: deadlock resolution10}). 
Therefore, at least one agent does not have a blocking agent, ensuring that the proposed algorithm does not cause a deadlock.
\end{proof}

\subsection{Enhancing Waypoint Update Speed via Communication}
\label{subsec:update_speed}
The communication-free method (MC-Swarm-N) requires that, for waypoint updates, agents often wait until all agents’ subgoals have reached their waypoints, which may increase the overall flight time.
To mitigate this issue, Algorithm \ref{alg3:update_waypoints_communication} introduces a faster waypoint update method using minimal communication. Each agent locally identifies which agents' subgoals have reached the waypoint. The algorithm then updates the waypoint only for agents that all agents agree have reached their respective waypoints, whereas Algorithm \ref{alg2:update_waypoints} pauses the update until all subgoals of agents arrive at their waypoints.

To elaborate, each agent communicates a set of agent indices $\mathbf{e}_i$ indicating which agents it observes to have reached their respective waypoints. The intersection of these sets across all agents, $\mathbf{e}_s = \bigcap_{i \in \mathcal{I}} \mathbf{e}_i$, identifies the agents for which all peers agree that the subgoals have been reached (line 2 of Algorithm \ref{alg3:update_waypoints_communication}). Only the agents whose indices belong to $\mathbf{e}_s$ proceed to update their waypoints, while the remaining agents retain their current subgoals (lines 7–13 of Algorithm \ref{alg3:update_waypoints_communication}).\\
For example, suppose $\mathcal{I}=\{1,2,3\}$, and agent 1, 2, and 3 perceive that the subgoals of agents $\{1, 2, 3\}$, $\{1, 2\}$, and $\{2, 3\}$ have arrived at their waypoints, respectively: $\textbf{e}_{1}=\{1, 2, 3\}$, $\textbf{e}_{2}=\{1, 2\}$, and $\textbf{e}_{3}=\{2, 3\}$.
In this case, only the waypoint of agent 2 is updated since all agents agree that the subgoal of agent 2 has reached its waypoint: $\textbf{e}_{s}=\{2\}$.
This approach maintains consistency across agents while significantly reducing communication overhead. Moreover, this accelerates the waypoint update speed because agents do not need to wait until all agents' subgoals reach their waypoint, reducing the total flight time. The planner that uses Algorithm \ref{alg3:update_waypoints_communication} instead of Algorithm \ref{alg2:update_waypoints} is referred to as MC-Swarm-C.


\subsection{Estimation Error Handling}
\label{subsec:error_handling}
Our system requires all agents to detect the positions of other agents, which makes the proposed algorithm naturally subject to state estimation errors. To mitigate the possible negative effects on safety due to the errors, we incorporated an additional safety margin into the agent's collision model. This safety margin is determined based on the maximum tracking error observed during preliminary experiments measuring the quadrotor's tracking error. Specifically, in the validation, we employ quadrotors with a 7.5 cm radius, corresponding to the size of Crazyflie 2.1 quadrotors, but their radii are regarded as 15 cm in the algorithm.

\begin{algorithm}[t!]
 \SetAlgoLined
\KwIn{Prev. waypoints $\textbf{w}^{(h-1)}_{\forall i \in \mathcal{I}}$, prev. subgoals $\textbf{g}^{(h-1)}_{\forall i \in \mathcal{I}}$, $\textbf{e}_{\forall i \in \mathcal{I}}$, and desired goal $\textbf{z}_{\forall i \in \mathcal{I}}$}
\KwOut{Waypoint for current state update step $\textbf{w}_{\forall i \in \mathcal{I}}$}
   \tcp{Grid-based MAPF}
   \textbf{do} lines 1-4 in Algorithm 2  

   $\textbf{e}_{s} \gets$ CheckMutualIndex($\textbf{e}_{\forall i \in \mathcal{I}}$)

   \tcp{Waypoint update (modified)}
   \uIf{h=0}{$\textbf{w}_{\forall i \in \mathcal{I}} \gets$ second waypoint of $\boldsymbol{\pi}_{\forall i \in \mathcal{I}}$\; 

   \textbf{goto} line 14
   }
   \textbf{end}

 \For{$j=1$ to $N$}{
    \uIf{$h=0$ or $j\in$ $\textbf{e}_{s}$}{$\textbf{w}_{j} \gets$ second waypoint of $\boldsymbol{\pi}_{j}$\;}
     \Else{$\textbf{w}_{j} \gets \textbf{w}^{(h-1)}_{j}$\;}
 }
   \tcp{Conflict resolution}
   \textbf{do} lines 10-17 in Algorithm 2.

   \KwRet{$\textbf{w}_{\forall i \in \mathcal{I}}$}
 \caption{Waypoints update (communication ver.)}
 \label{alg3:update_waypoints_communication}
 \end{algorithm}

\section{Validation}
\label{sec6:validation}
In this section, the proposed method is validated through both simulations and hardware experiments.
For simulations, tests are run with 30 different scenarios for each number of agents, varying the map configuration and the agents' initial positions. 
We then measure the success rate, total flight time, and computation time per agent. 
The performance of the proposed algorithms are investigated by comparing it with the following state-of-the-art asynchronous MATP algorithms.
\begin{itemize}
  \item MADER \cite{kondo2023robust}: Collision check-recheck-based
  \item EGO-v2 \cite{zhou2022swarm}: Gradient-based optimization-based
  \item DREAM \cite{csenbacslar2024dream}: Separating hyper-plane trajectory-based
  \item GCBF+ \cite{zhang2025gcbf+}: Graph neural network-based
\end{itemize}
We evaluate the performance of both proposed methods, described as follows:
\begin{itemize}
    \item MC-Swarm-N: Communication-free-based (waypoint update using Algorithm \ref{alg2:update_waypoints}),
    \item MC-Swarm-C: Commnucation-based (waypoint update using Algorithm \ref{alg3:update_waypoints_communication}).
\end{itemize}

For parameters, we set the collision model of the agent with radius $r_{\forall i} = 0.15$ m, and the maximum axis-wise velocity and acceleration of the agents are $v_{\max} = 1.0$ m/s and $a_{\max} = 5.0$ m/$\text{s}^2$, respectively. The update parameters, such as the state update period, maximum replanning period, and planning period, used in validation are set as $T_{s} = 0.02\  \text{s}$, $T_{r} = 0.2\  \text{s}$, and $\Delta t_{h} = 0.1\  \text{s}$. For the MAPF algorithm, the grid resolution $d$ is set to $0.5$ m.
For trajectory optimization, we set the planning horizon $T=1.0$ s, with the number of steps $M=5$ and time step $\Delta t=0.2$ s, and the weight parameters in the cost (\ref{eq: cost function}) are set as $w_{e}=0.01$ and $w_{g}=0.1$. In the implementation, Octomap \cite{hornung2013octomap} is used to represent the obstacle environment, and the IBM CPLEX QP solver \cite{cplex201612} is used for trajectory optimization. We use a laptop with an Intel i7-10750H CPU for simulations and an Intel NUC with an i7-1260P CPU for experiments, respectively.
\begin{figure}
    \centering
    \includegraphics[width=0.5\linewidth]{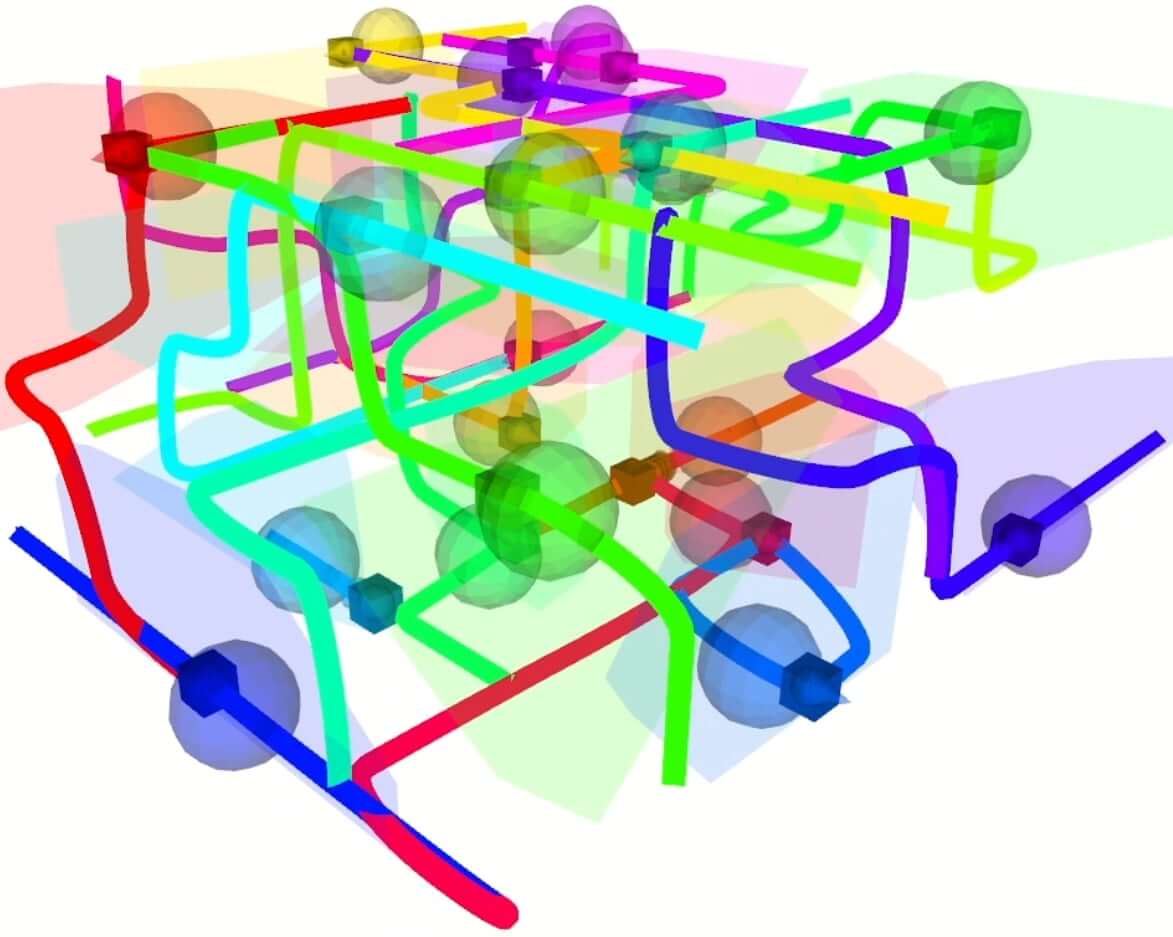}
    \caption{Simulation result in empty space. Spheres, splines, and shaded regions represent agents, whole flying paths, and collision-free regions, respectively.}
    \label{fig6:empty_space_simulation}
\end{figure}
\begin{table}[t!]
    \centering
    \caption{Simulation results in empty spaces.}
    \label{tab6:sim_empty_results}
\scalebox{0.95}{
\begin{tabular}{cc|ccc}
\hline
$N$&Method & Success rate [\%] &Flight time [s] &Runtime [ms]\\ 
\hline
\multirow{3}{*}{5}&DREAM & 100 & 4.77 & 2.59 \\
&MC-Swarm-N & \textbf{100} & \textbf{4.33} & 2.27 \\ 
&MC-Swarm-C & \textbf{100} & 4.82 & 2.25 \\ 
\hline
\multirow{3}{*}{10}&DREAM & 100 & 6.76 & 5.30 \\
&MC-Swarm-N & \textbf{100} & 7.12 & 3.73 \\ 
&MC-Swarm-C & \textbf{100} & \textbf{6.52} & 3.43 \\ 
\hline
\multirow{3}{*}{15}&DREAM & 93.3 & 11.3 & 22.5 \\
&MC-Swarm-N & \textbf{100} & 8.63 & 5.07 \\ 
&MC-Swarm-C & \textbf{100} & \textbf{7.36} & 5.63 \\ 
\hline
\multirow{3}{*}{20}&DREAM & 70.0 & 12.79 & 34.1 \\
&MC-Swarm-N & \textbf{100} & 10.8 & 7.23 \\ 
&MC-Swarm-C & \textbf{100} & \textbf{7.97} & 9.62 \\ 
\hline
\end{tabular}}
\footnotesize{The bold number indicates the best result.\quad \quad \quad \quad \quad \quad \quad \quad \quad \quad \quad \quad \quad \ \ } 
\end{table}
\subsection{Simulation in Obstacle-Free Space}
\label{subsec:simul_obstacle_free}
We conduct the simulations with 5 to 20 agents in a $3\times3\times1$ $\text{m}^{3}$ obstacle-free space. We execute 30 trials for each number of agents, randomly assigning start and goal points for each test. 
Fig. \ref{fig6:empty_space_simulation} shows the simulation result carried out by MC-Swarm-N, and Table \ref{tab6:sim_empty_results} summarizes the result.
Our methods successfully accomplish the mission, whereas DREAM fails to complete it due to a deadlock that occurs as the number of agents increases. Additionally, although MC-Swarm-C takes more time than MC-Swarm-N to complete the mission with 5 agents, it becomes faster as the number of agents increases, with the time difference growing more significant.
\begin{figure}[t!]
    \centering
    \begin{subfigure}[t]{0.241\textwidth}
    \centering \includegraphics[width=1.0\linewidth]{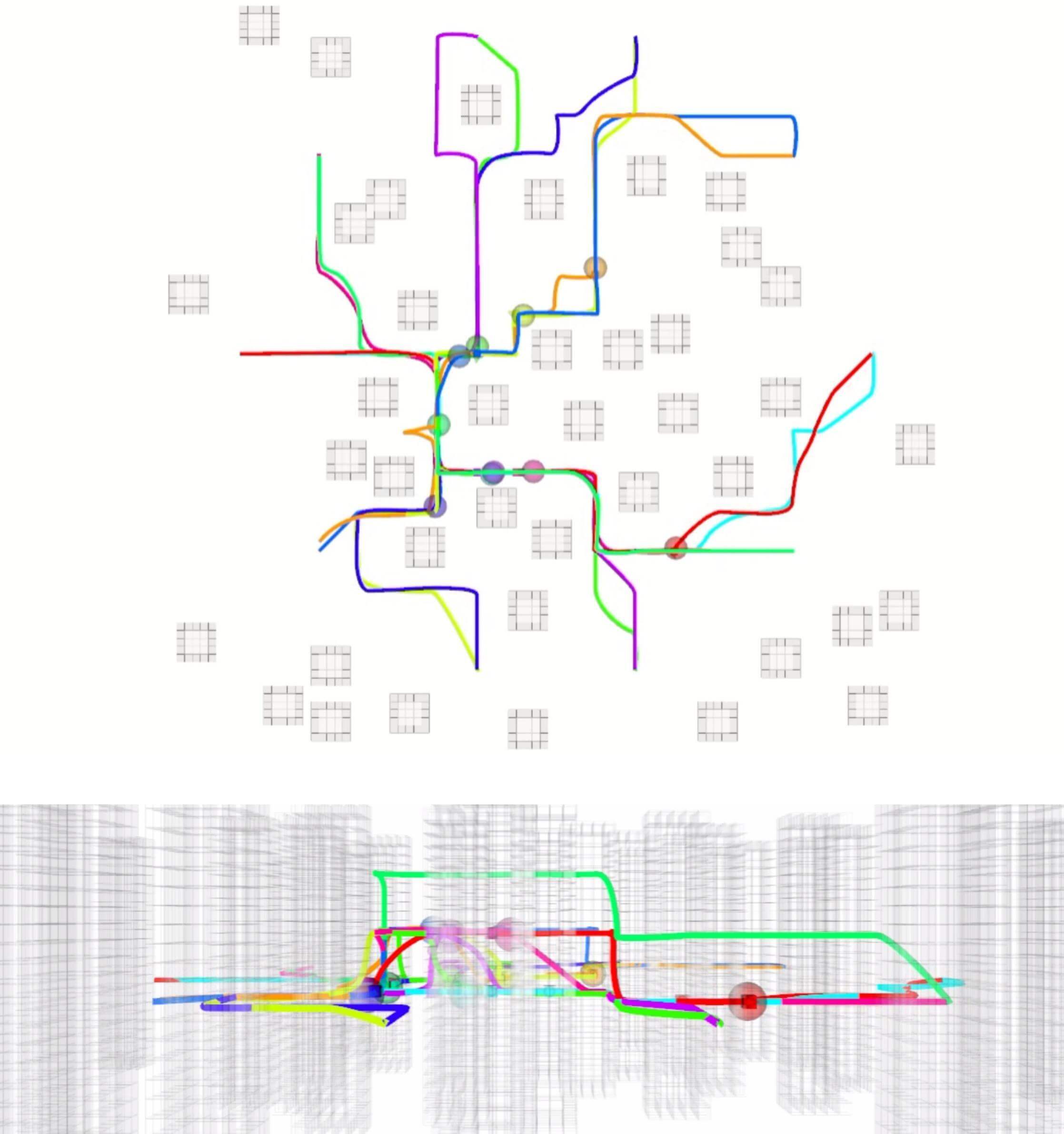}
    \caption{Random forest}
    \label{fig6:sim_forest_3d}
    \end{subfigure}
    \begin{subfigure}[t]{0.241\textwidth}
    \centering \includegraphics[width=1.0\linewidth]{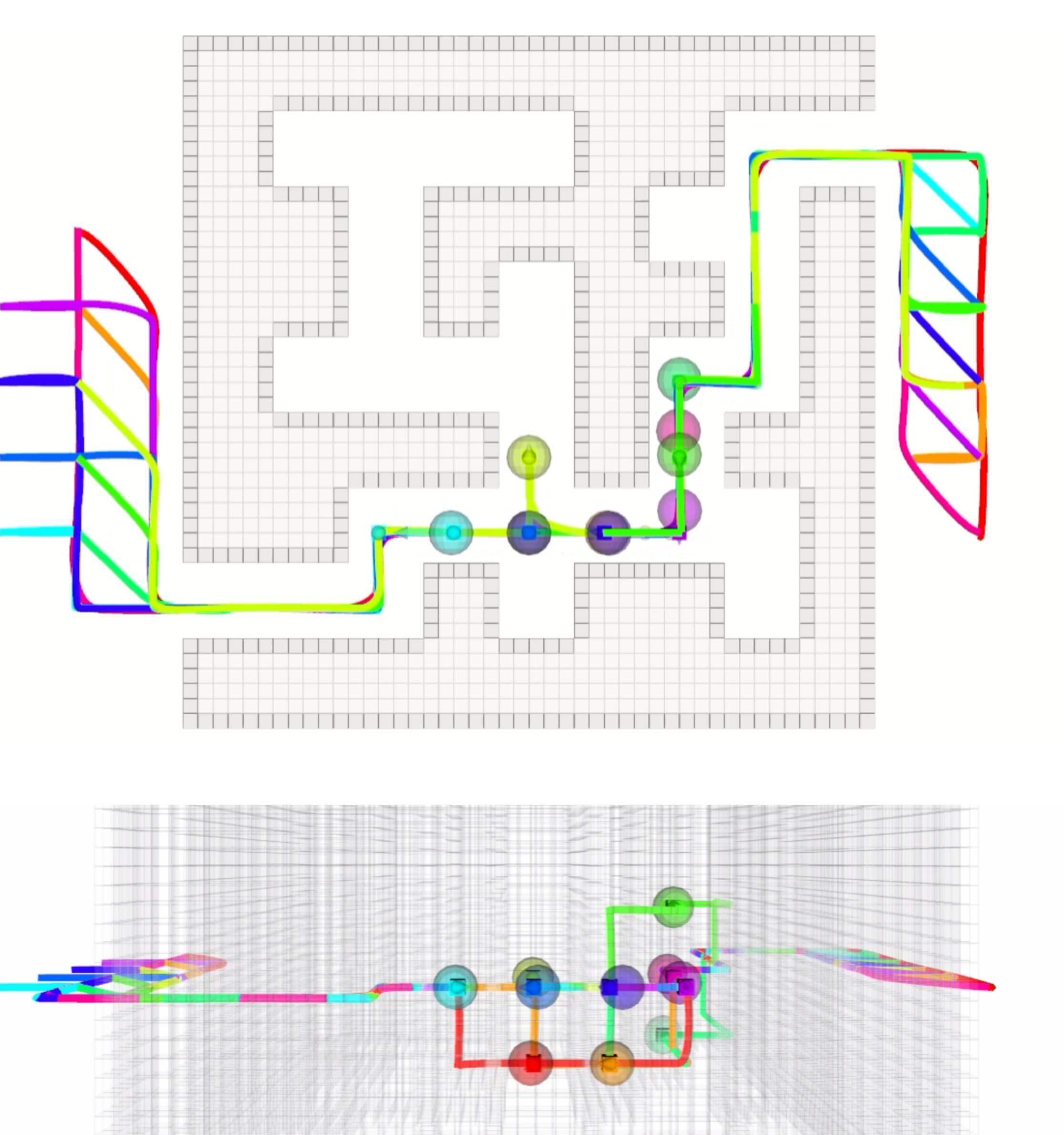}
    \caption{Narrow maze}
    \label{fig6:sim_maze_3d}
    \end{subfigure}
    \caption{Simulations in 3D obstacle environments using MC-Swarm-N (top: top-down view, bottom: side view). Splines and spheres show the reported trajectories and agents, respectively. }
    \label{fig6:sim_3d}
\end{figure}
\begin{figure}[t!]
    \centering
    \begin{subfigure}[t]{0.241\textwidth}
    \centering \includegraphics[width=1.0\linewidth]{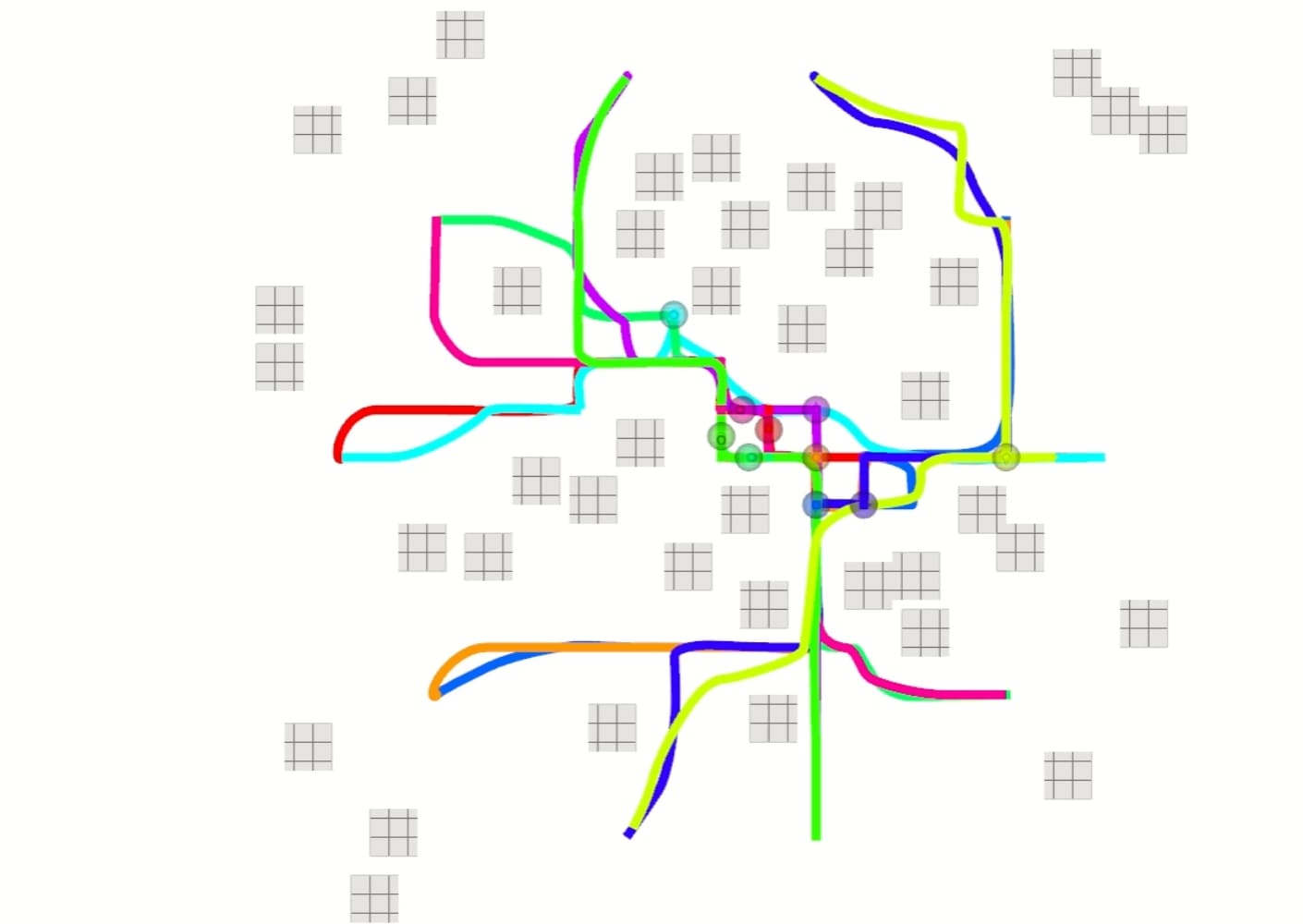}
    \caption{Random forest}
    \label{fig6:sim_forest_2d}
    \end{subfigure}
    \begin{subfigure}[t]{0.241\textwidth}
    \centering \includegraphics[width=1.0\linewidth]{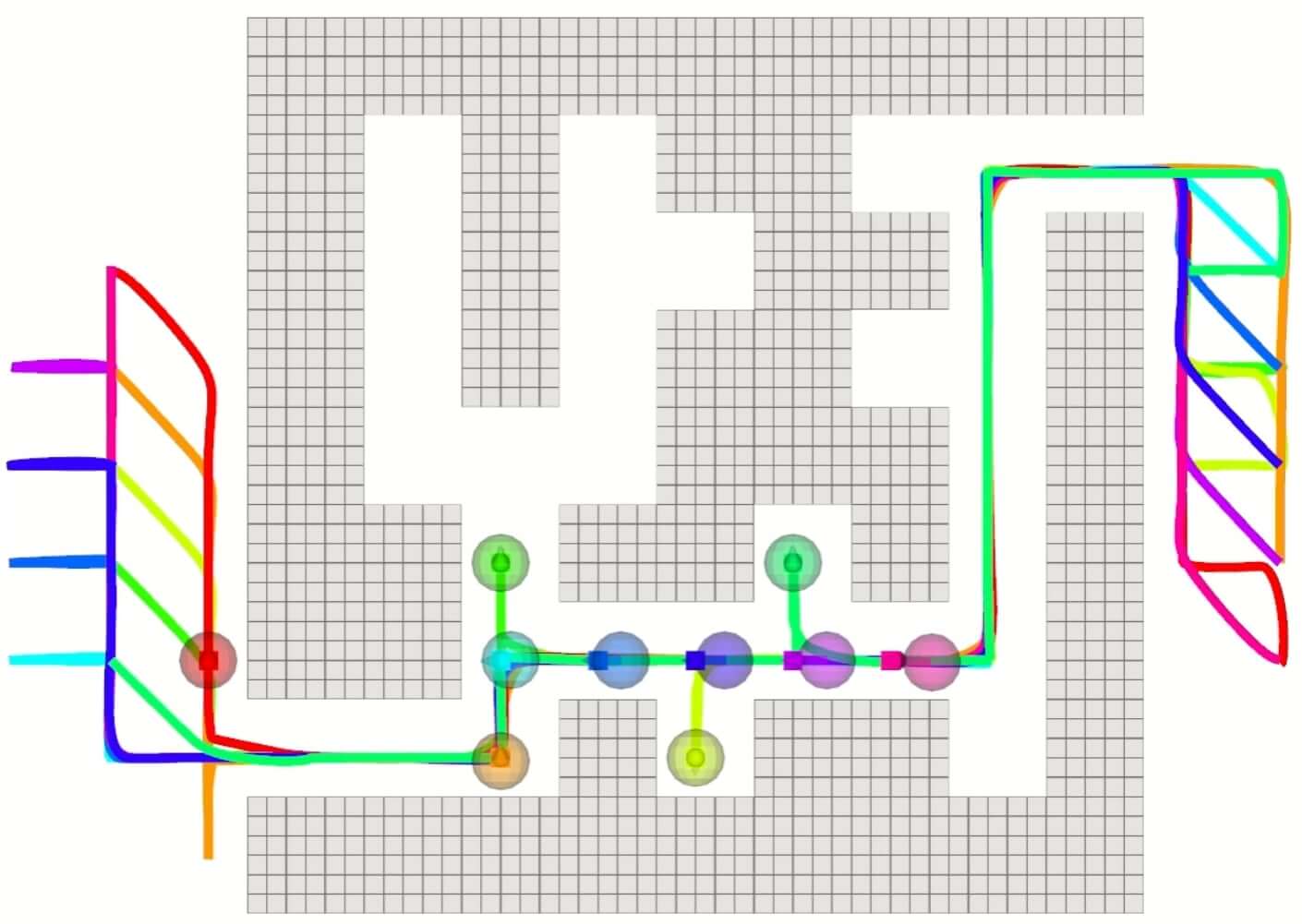}
    \caption{Narrow maze}
    \label{fig6:sim_maze_2d}
    \end{subfigure}
    \caption{Simulations in 2D obstacle environments using MC-Swarm-N. Splines and spheres show the reported trajectories and agents, respectively.}
    \label{fig6:sim_2d}
\end{figure}
\subsection{Simulation in Obstacle Space}
\label{subsec:simul_obstacle_yes}
We perform goal-reaching simulations in two types of obstacle environments. First, we test our algorithm in random forests. Forty box-shaped obstacles are deployed in random positions, and all agents are on a circle with $4$ m radius at initial time, as shown in Fig. \ref{fig6:sim_forest_3d}. Then, all agents are commanded to move to their goal points, which are antipodal to their start points. Second, we conduct the test in a narrow-gap maze. With a maze environment in the center, all agents start from the left and right sides and exit through the opposite side, as shown in Fig. \ref{fig6:sim_maze_3d}. There is only one entrance on each side of the maze, and the path to their destinations is unique. Only a single agent can pass through the corridor at a time.
Fig. \ref{fig6:sim_3d} illustrates the simulation results conducted using MC-Swarm-N, while Table \ref{tab6:sim3d_results} provides a summary of the results. GCBF+ failed to reach the goal in both environments as the agents got stuck between obstacles. While MADER and EGO-Swarm succeeded in some cases in the forest environment with short total flight time, they failed in all cases in the maze environments, which are significantly denser. On the other hand, the proposed methods achieve a 100\% success rate by resolving deadlocks.
\begin{table}[t!]
    \centering
    \caption{Simulation results in 3D obstacle spaces.} 
    \label{tab6:sim3d_results}
\scalebox{0.96}{\begin{tabular}{cc|ccc}
\hline
&Method & Success rate [\%] & Flight time [s] & Runtime [ms]\\ 
\hline
\multirow{5}{*}{{\rotatebox[origin=c]{90}{Forest}}}&GCBF+ & 0.00 & - & - \\ 
&MADER & 43.3 & 19.5 & 226 \\
&EGO-v2 & 83.3 & \textbf{14.5} & 5.19 \\
&MC-Swarm-N & \textbf{100} & 26.9 & 2.61 \\ 
&MC-Swarm-C & \textbf{100} & 19.7 & 2.81 \\ 
\hline
\multirow{5}{*}{{\rotatebox[origin=c]{90}{Maze}}}&GCBF+ & 0.00 & - & -\\ 
&MADER & 0.00 & - & - \\
&EGO-v2 & 0.00 & - & - \\
&MC-Swarm-N& \textbf{100} & 65.9 & 3.04 \\ 
&MC-Swarm-C & \textbf{100} & \textbf{36.7} & 2.86 \\ 
\hline
\end{tabular}}
\footnotesize{The bold number indicates the best result. \quad \quad \quad \quad \quad \quad \quad \quad \quad \quad \quad \quad \ \ }
\end{table}
\begin{table}[t!]
    \centering
    \caption{Simulation results in 2D obstacle spaces.} 
    \label{tab6:sim2d_results}
\scalebox{0.96}{\begin{tabular}{cc|ccc}
\hline
&Method & Success rate [\%] & Flight time [s] & Runtime [ms]\\ 
\hline
\multirow{3}{*}{{\rotatebox[origin=c]{90}{Forest}}}&GCBF+& 0.00 & - & -\\ 
&MC-Swarm-N & \textbf{100} & 33.3 & 2.00 \\ 
&MC-Swarm-C & \textbf{100} & \textbf{23.5} & 2.08 \\ 
\hline
\multirow{3}{*}{{\rotatebox[origin=c]{90}{Maze}}}&GCBF+ & 0.00 & - & -\\ 
&MC-Swarm-N & \textbf{100} & 99.9 & 3.17 \\ 
&MC-Swarm-C & \textbf{100} & \textbf{63.0} & 2.14 \\ 
\hline
\end{tabular}}
\footnotesize{The bold number indicates the best result. \quad \quad \quad \quad \quad \quad \quad \quad \quad \quad \quad \quad \ \ }
\end{table}
\begin{figure}[t!]
    \centering
    \includegraphics[width = 1.0\linewidth]{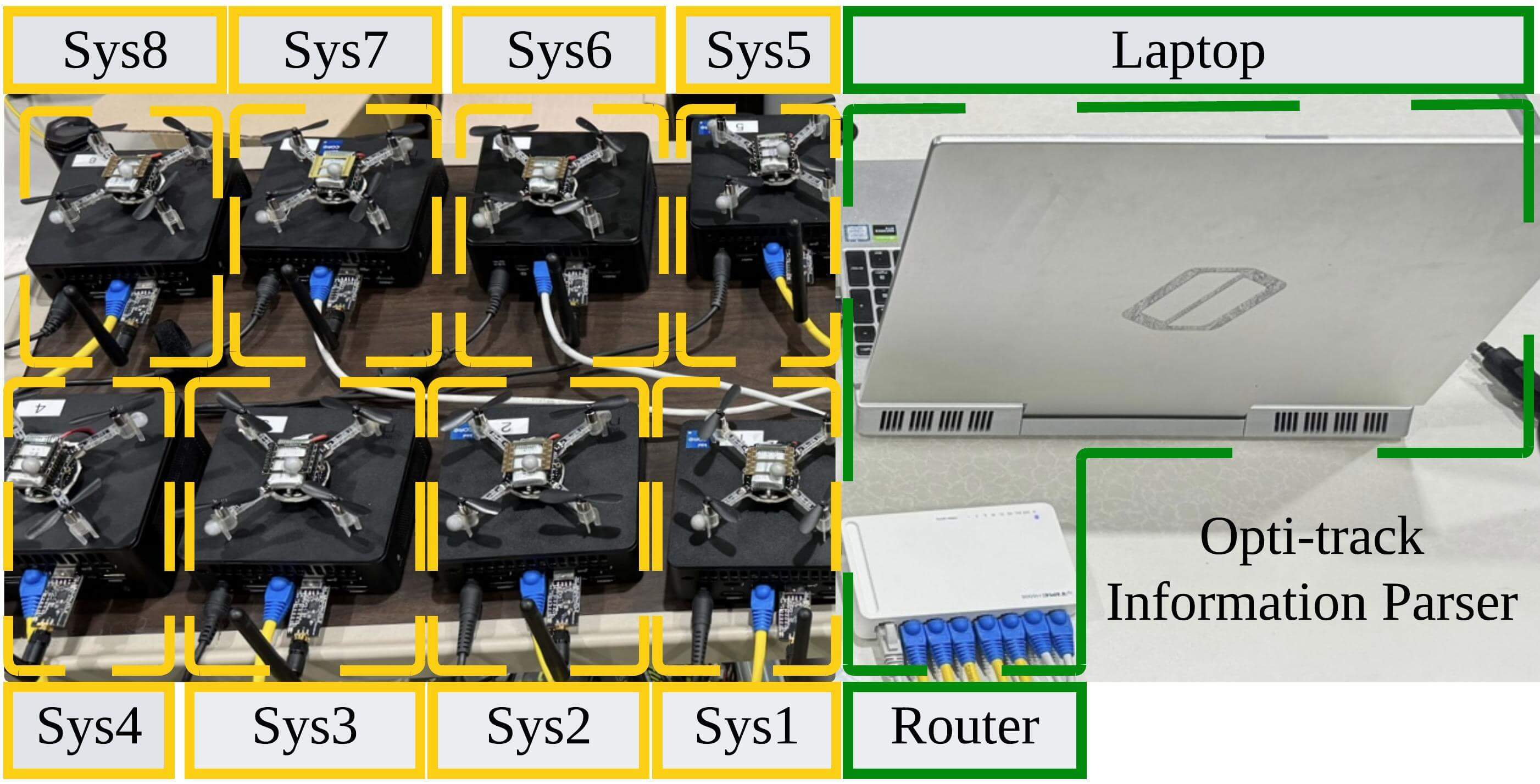}
    \caption{Hardware demonstration setup.
    }
    \label{fig6:exp_settings}
\end{figure}

Additionally, we test the planner in 2D spaces with the same obstacle layout as in the 3D spaces, making the scenarios more challenging than the 3D cases, as feasible motions must be found within the x–y plane. Fig. \ref{fig6:sim_2d} and Table \ref{tab6:sim2d_results} summarize the simulation results, and only our methods succeeds in the tasks without any failure.
\subsection{Experiments}
\label{subsec:val_experiments}
For hardware demonstration, we set up a validation system, as shown in Fig. \ref{fig6:exp_settings}. Eight Crazyflie \cite{crazyflie} quadrotors are employed, each of which communicates with and is controlled by a separate Intel NUC computer. Also, an additional laptop is used to receive data from the Optitrack motion capture system, and the positional data of the quadrotors is asynchronously transferred to the computers via a router.
\begin{figure}[t!]
    \centering
    \begin{subfigure}[t!]{0.262\textwidth}
        \centering
        \includegraphics[width=\textwidth]{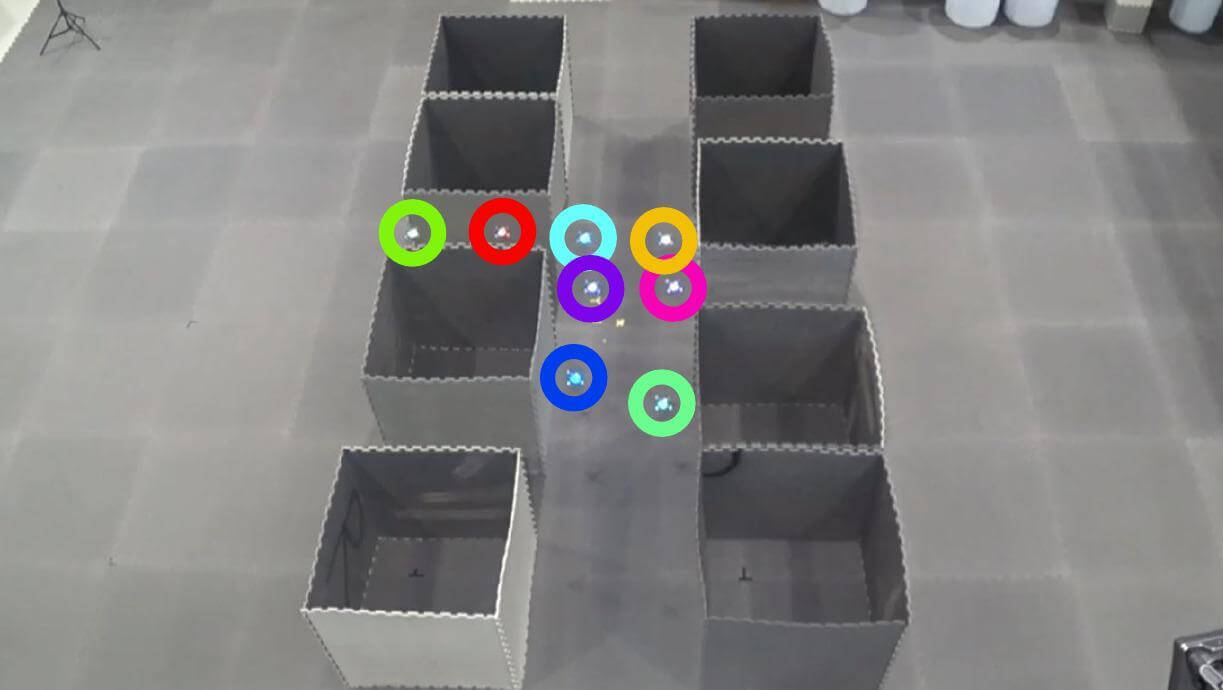}
        \caption{Third person view ($t = 10\ \text{s}$)}
        \label{subfig6:dr_exp_cam}
    \end{subfigure}
    \begin{subfigure}[t!]{0.217\textwidth}
        \centering
        \includegraphics[width=\textwidth]{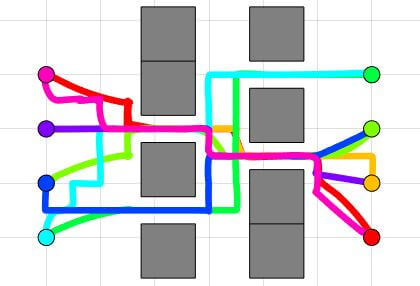}
        \caption{Flight history, ($t = 38.2\  \text{s}$)}
        \label{subfig6:dr_exp_total}
    \end{subfigure}
    \caption{Real-world experiments using the no-communication method (MC-Swarm-N) with eight quadrotors. (a): Snapshots of flight at $t = 10\ \text{s}$. (b): Reported flight paths and snapshots at the end time.}
    \label{fig6:abvc_dr_exp}
\end{figure}
\begin{figure}[t!]
    \centering
    \begin{subfigure}[t!]{0.262\textwidth}
        \centering
        \includegraphics[width=\textwidth]{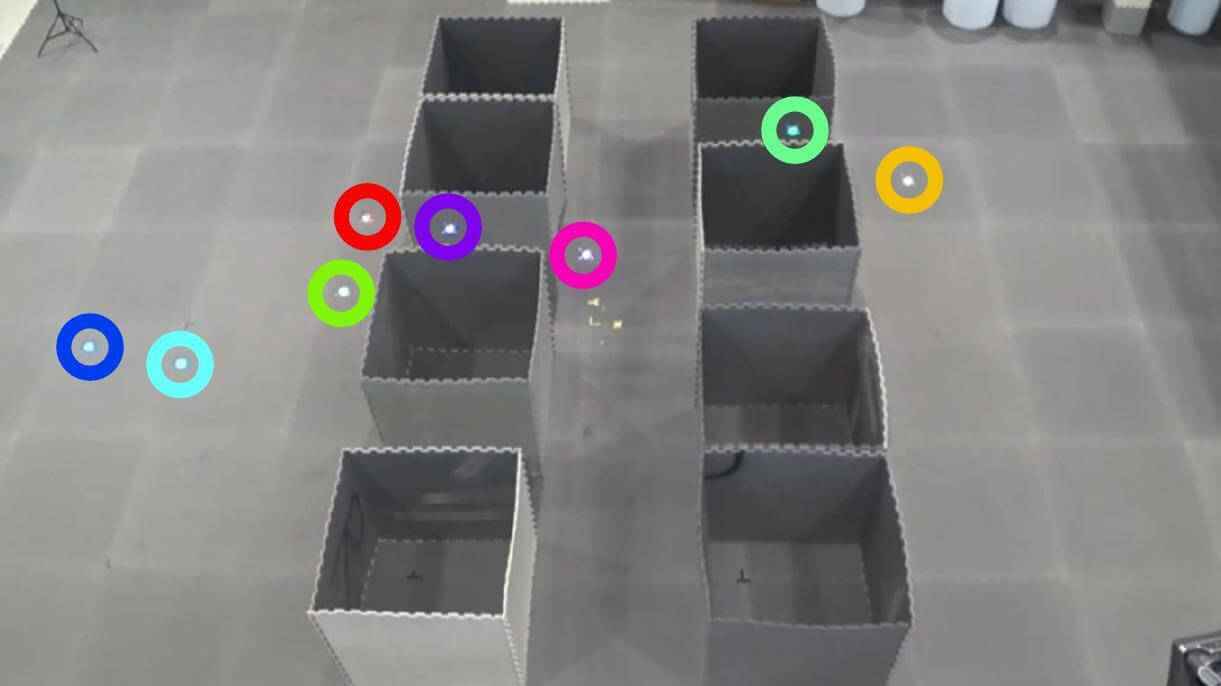}
        \caption{Third person view ($t = 10\ \text{s}$)}
        \label{subfig6:drc_exp_cam}
    \end{subfigure}
    \begin{subfigure}[t!]{0.219\textwidth}
        \centering
        \includegraphics[width=\textwidth]{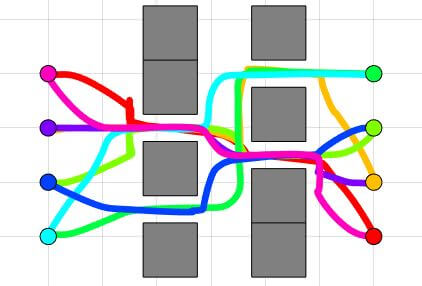}
        \caption{Flight history, ($t = 24.9\  \text{s}$)}
        \label{subfig6:drc_exp_total}
    \end{subfigure}
    \caption{Real-world experiments using the communication-based (MC-Swarm-C) with eight quadrotors. (a): Snapshots of flight at $t = 10\ \text{s}$. (b): Reported flight paths and snapshots at the end time.}
    \label{fig6:abvc_drc_exp}
\end{figure}

In the demonstration, we allow the drones to move within an $8\times5\ \text{m}^{2}$ 2D space, where eight $1\times1\ \text{m}^{2}$ box-shaped obstacles are arranged to form a narrow-gap maze.
Each agent is assigned to reach a goal point that is symmetric to the start point with respect to the origin, passing through a $0.5\ \text{m} $ gap. Through experiments, we show successful goal-reaching scenarios, and Table \ref{tab6:exp_results}, Figs. \ref{fig6:abvc_dr_exp} and \ref{fig6:abvc_drc_exp} summarize the flights.
Fig. \ref{fig6:abvc_dr_exp} shows the result with MC-Swarm-N, while Fig. \ref{fig6:abvc_drc_exp} presents the result with MC-Swarm-C. Figs. \ref{subfig6:dr_exp_cam} and \ref{subfig6:drc_exp_cam} show snapshots 10 seconds after the start of the flight.
 Unlike the communication-free method, the communication-based method's faster update speed allows the agents to exit the confined space earlier and move closer to the destination at the same time. Ultimately, the communication-based method requires communication but leads to completing the mission more quickly, as shown in Table \ref{tab6:exp_results}.

\begin{table}[t]
    \centering
    \caption{Reported results in hardware experiments} 
    \label{tab6:exp_results}
\begin{tabular}{c|cc}
\hline
Method & MC-Swarm-N  & MC-Swarm-C\\ 
\hline
Flight time [s] & 38.2 & \textbf{24.9} \\ 
\hline
Runtime [ms] & 2.18 & 2.25\\
\hline
\end{tabular}\\
\footnotesize{The bold number indicates the best result. \quad \ \ }
\end{table}

\section{Conclusion}
\label{sec7:conclusion}
We presented the asynchronous and distributed trajectory planning algorithm for quadrotor swarms. By ensuring that the agents have the same coordination states, such as waypoints, subgoals, and safety constraints, we enabled the quadrotor swarm to operate asynchronously without communication between the agents. We proved that the proposed algorithm guarantees 1) collision avoidance against the other agents and static obstacles and 2) deadlock resolution. Additionally, to improve the slow operation speed, we proposed a method that relies on communication but enables fast coordination state updates, thereby reducing the overall flight time.
We confirmed that the proposed methods achieve a 100\% success rate in environments with densely located obstacles, while the state-of-the-art (SOTA) algorithms show poor performance. Lastly, we conducted the challenging hardware demonstration in the narrow-gap environments with eight Crazyflie quadrotors, and there was no collision and deadlock during the flight.


\bibliographystyle{./bibtex/IEEEtran}
\bibliography{./bibtex/IEEEabrv, ./bibtex/mybibfile}

\begin{IEEEbiography}[{\includegraphics[width=1in,height=1.25in,clip,keepaspectratio]{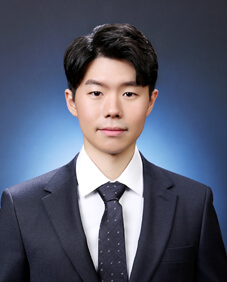}}]{Yunwoo Lee}
Yunwoo Lee received his B.S. degree in Electrical and Computer Engineering and his Ph.D. degree in Mechanical and Aerospace Engineering from Seoul National University, Seoul, South Korea, in 2019 and 2025, respectively. He is currently working for the AI Institute of Seoul National University and conducting on-site research at Carnegie Mellon University. His current research interests include multi-robot systems and vision-based trajectory planning for MAVs.
\end{IEEEbiography}

\begin{IEEEbiography}[{\includegraphics[width=1in,height=1.25in,clip,keepaspectratio]{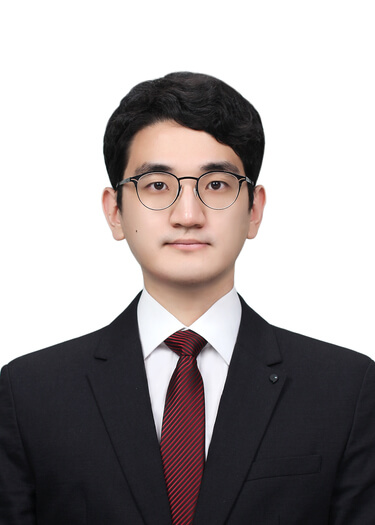}}]{Jungwon Park}
received the B.S. degree in electrical and computer engineering in 2018, and the M.S and Ph.D. degrees in mechanical and aerospace engineering at Seoul National University, Seoul, South Korea in 2020 and 2023, respectively. He is currently an assistant professor at Seoul National University of Science and Technology (SeoulTech), Seoul, South Korea. His current research interests include path planning and task allocation for distributed multi-robot systems. His work was a finalist for the Best Paper Award in Multi-Robot Systems at ICRA 2020 and won the top prize at the 2022 KAI Aerospace Paper Award.
\end{IEEEbiography}



\vspace{11pt}

\vfill

\end{document}